\pdfoutput=1

\documentclass[11pt]{article}

\usepackage[final]{acl}

\usepackage{times}
\usepackage{latexsym}

\usepackage[T1]{fontenc}

\usepackage[utf8]{inputenc}

\usepackage{microtype}

\usepackage{inconsolata}
\usepackage{CJKutf8}
\usepackage{ulem}

\definecolor{mypurple1}{RGB}{143, 94, 255}
\definecolor{myblue1}{RGB}{59, 76, 206}
\definecolor{myred1}{RGB}{190, 90, 75}
\definecolor{myyellow1}{RGB}{255, 192, 0}
\definecolor{mygreen1}{RGB}{160, 194, 128}
\newcommand{\multilinecolortext}[2]{\color{#1}{#2}}

\newcommand{\highlight}[1]{\textbf{#1}}
\newcommand{\highlighttwo}[1]{\uline{#1}}
\usepackage{booktabs}
\usepackage{graphicx}
\usepackage{xcolor}
\usepackage{multicol}
\usepackage{multirow}
\usepackage{paracol}
\usepackage{caption}
\usepackage[many]{tcolorbox}
\usepackage{cuted}
\usepackage{subcaption}
\usepackage{makecell} 
\usepackage{float}
\usepackage{xspace}
\usepackage{enumitem}
\newcommand{\name}{MT-Eval\xspace}
\NewTColorBox{NewBox}{ s O{!htbp} }{%
  floatplacement={#2},
  IfBooleanTF={#1}{float*,width=\textwidth}{float},
  breakable
  }

\newtcolorbox[blend into=figures]{prompt}[2][]{enhanced,
  title={#2},
  left=2pt,right=2pt,top=0pt,bottom=0pt,
  colframe=black,colback=white,colbacktitle=white,
  fonttitle=\bfseries,coltitle=black,
  minipage boxed title*=-3cm,
  attach boxed title to bottom center,
  floatplacement=!htbp,
  #1}
\title{{\name}: A Multi-Turn Capabilities Evaluation Benchmark for \\Large Language Models}

\author{Wai-Chung Kwan$^1$\Thanks{ Work done during an internship at Huawei Noah's Ark Lab.}, Xingshan Zeng$^2$, Yuxin Jiang$^3$,Yufei Wang$^2$, Liangyou Li$^2$, \\
\textbf{Lifeng Shang$^2$, Xin Jiang$^2$,Qun Liu$^2$, Kam-Fai Wong$^1$}  \\
$^1$The Chinese University of Hong Kong \enspace
$^2$Huawei Noah's Ark Lab \\
$^3$The Hong Kong University of Science and Technology \\
\{wckwan,kfwong\}@se.cuhk.edu.hk \\ \{zeng.xingshan,wangyufei44,liliangyou,Shang.Lifeng,jiang.xin,qun.liu\}@huawei.com \\ yjiangcm@connect.ust.hk
}

\begin{document}
\maketitle
\begin{abstract}
  Large language models (LLMs) are increasingly relied upon for complex multi-turn conversations across diverse real-world applications.
  However, existing benchmarks predominantly focus on single-turn evaluations, overlooking the models' capabilities in multi-turn interactions.
  To address this gap, we introduce {\name}, a comprehensive benchmark designed to evaluate multi-turn conversational abilities.
  By analyzing human-LLM conversations, we categorize interaction patterns into four types: recollection, expansion, refinement, and follow-up.
  We construct multi-turn queries for each category either by augmenting existing datasets or by creating new examples with GPT-4 to avoid data leakage.
  To study the factors impacting multi-turn abilities, we create single-turn versions of the 1170 multi-turn queries and compare performance.
  Our evaluation of 11 well-known LLMs shows that while closed-source models generally surpass open-source ones, certain open-source models exceed GPT-3.5-Turbo in specific tasks.
  We observe significant performance degradation in multi-turn settings compared to single-turn settings in most models, which is not correlated with the models' fundamental capabilities.
  Moreover, we identify the distance to relevant content and susceptibility to error propagation as the key factors influencing multi-turn performance.
    {\name} is released publicly to encourage future research towards more robust conversational models \footnote{Code and data are available at \url{https://github.com/KwanWaiChung/MT-Eval}}.
\end{abstract}

\section{Introduction}

The rise of large language models (LLMs) is transforming our daily lives and professional endeavors with their growing capabilities.
Individuals increasingly rely on LLM-based AI assistants for diverse tasks, such as coding assistance, summarizing text from documents, and devising business strategies \citep{zheng_lmsys-chat-1m_2023}.
These tasks often require understanding complex instructions and adapting to evolving needs through multiple user interactions.
Moreover, it is crucial for LLMs to generate contextually coherent responses by retaining and recalling historical information.
The ability of LLMs to engage in multi-turn conversations is often overlooked in existing evaluation frameworks.
For instance, MMLU \citep{hendrycks_measuring_2020} evaluates language understanding in multiple tasks using single queries, and MT-Bench \citep{zheng_judging_2023} evaluates conversational ability using two-turn interactions without considering more turns and various conversation types.

\begin{figure*}[!htbp]
  \centering
  \includegraphics[width=\textwidth]{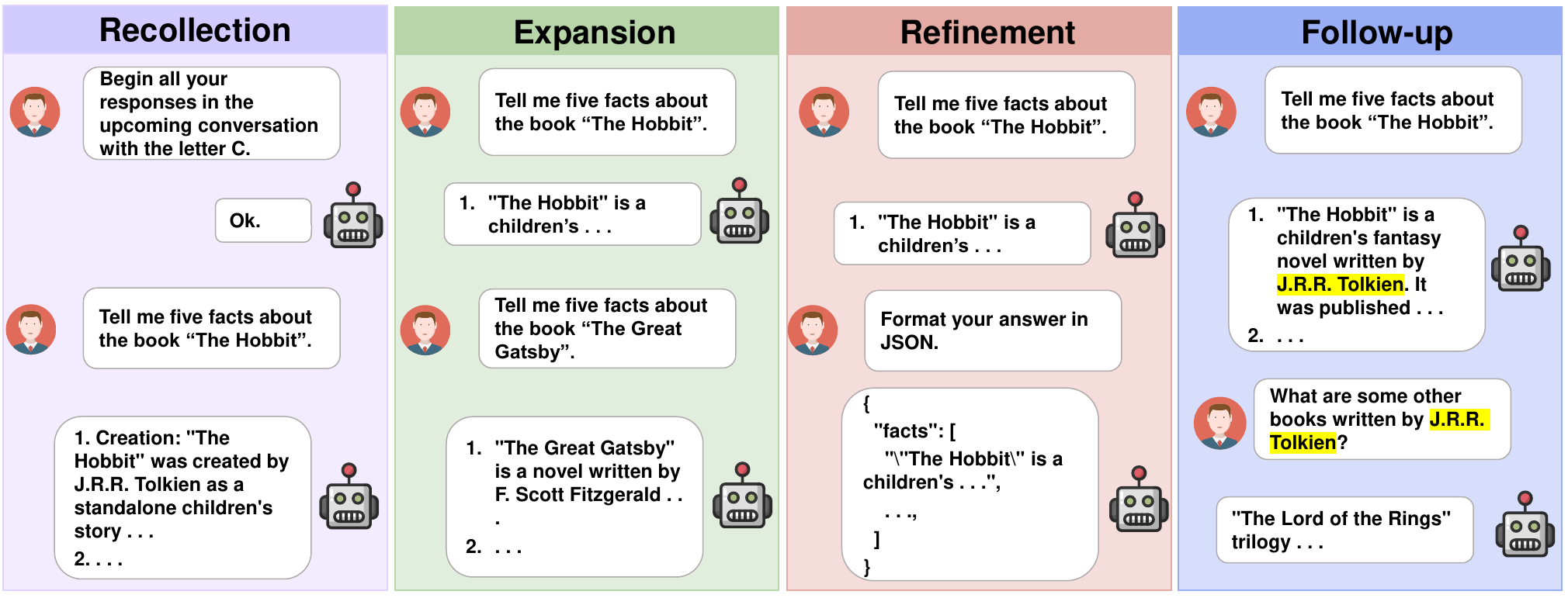}
  \caption{Illustration of the four dialogue tasks in {\name}: Recollection, Expansion, Refinement, and Follow-up.
    Recollection accesses the model's ability to recall information from previous conversations.
    Expansion evaluates the model's capacity to address queries surrounding the same topic.
    Refinement gauges the model's adherence to progressively complex instructions.
    Follow-up examines the model's proficiency in responding to queries that build upon its preceding response. A more detailed description of these tasks can be found in Section \ref{sec:benchmark}.}
  \label{fig:four-task-examples}

\end{figure*}

To bridge the gap, we propose \name, an evaluation benchmark to measure the capabilities of LLMs to conduct coherent multi-turn conversations.
Our analysis of interactions in LMSYS-Chat-1M \citep{zheng_lmsys-chat-1m_2023} reveals four predominant patterns when users interact with AI assistants:
Recollection, where the assistant must recall information from earlier turns; Expansion, involving the exploration of varied topics within the main subject; Refinement, where initial instructions are clarified or revised; and Follow-up, consisting of questions based on the assistant's previous responses (see Figure~\ref{fig:four-task-examples}).
These patterns are reflective of the majority of real-world multi-turn interactions with assistants.
We then construct evaluation sets for each interaction type by augmenting existing datasets or creating new ones to cover real-world applications.
We use GPT-4 to generate new instances to avoid data contamination, and we manually review and revise them for quality assurance.

A performant multi-turn conversational model should perform well in multi-turn interactions and demonstrate minimal performance difference from the corresponding single-turn scenario.
Therefore, we compare models using both response quality in the multi-turn setting and performance differences for identical queries in single vs. multi-turn settings.
Combining these two results provides a comprehensive view of their multi-turn conversational capabilities.

We evaluate 11 popular LLMs, including both open-source and closed-source models.
Beyond general evaluation, we conduct in-depth analysis and ablation studies revealing how LLMs conduct multi-turn interactions and what affects their performance. Our findings include:
1) GPT-4 still dominates in multi-turn conversational abilities, but some open-source models have comparable performance to GPT-3.5-Turbo in some tasks.
2) Most LLMs perform worse in the multi-turn setting than in single-turn. The performance gap between the two settings is not related to the model's fundamental capacities.
3) Increasing distance to relevant content negatively impacts performance.
4) Models are prone to error propagation due to sensitivity to dialogue history.

We summarize our contributions as follows:
\begin{itemize}[leftmargin=0.5cm, itemsep=0.01cm]
  \item We propose a comprehensive multi-turn conversational capabilities evaluation benchmark that covers a wide range of real-world scenarios.
  \item We provide an in-depth analysis of the performance of  11 popular LLMs across our benchmark, offering insights into their capabilities in multi-turn conversations.
  \item We identify key factors that influence LLM multi-turn performance, such as the distance to relevant content and error propagation.
  \item We demonstrate the importance of evaluating LLMs in multi-turn settings, highlighting the performance discrepancies that can arise when compared to single-turn evaluations.
\end{itemize}

\section{Related Work}
Recent advancements in LLMs \citep{openai_gpt-4_2023, touvron2023llama, vicuna2023} have significantly improved their ability to engage in human-like, multi-turn conversations. These models can now understand instructions, intentions, and context from human prompts, offering valuable responses. \citep{zhao2023survey}.
However, a limited number of studies have delved into the multi-turn conversation capabilities of LLMs.
\citet{zheng_judging_2023} developed MT-Bench, a dataset comprising 80 meticulously crafted multi-turn questions designed to evaluate the conversational flow and instruction-following capabilities of LLMs.
Nevertheless, the dataset's limited sample size poses a challenge, with each conversation consisting of only two turns.
This constraint hinders the ability to broaden the evaluation scope or capture the intricacies of more extended conversational contexts.
\citet{lee_evaluating_2023} proposed HALIE, a framework for evaluating human-AI interaction. but its reliance on human participation limits its scalability and efficiency across different tasks.
In specific domains, \citet{liao2023automatic} designed an automatic evaluation framework for multi-turn medical consultations capabilities of LLMs.
Moreover, \citet{wang_mint_2023} proposed MINT to evaluate LLMs' ability to solve tasks with multi-turn interactions.
It focuses on LLM's ability to use tools and utilize feedback during multi-turn conversations.
In contrast, our work evaluates LLM's comprehensive ability to conduct multi-turn conversations, possibly involving multiple types of dialogue in one session.

\begin{table}[ht]
  \small
  \centering
  \begin{tabular}{l|r}
    \toprule
    Statistics                          & Number \\
    \midrule
    Avg. \# Turns per Dialogue          & 6.96   \\
    Avg. \# Words in Prompt $\dagger$   & 760.41 \\
    Max. \# Words in Prompt $\dagger$   & 2574   \\
    Avg. \# Words in Response $\dagger$ & 99.31  \\
    Max. \# Words in Response $\dagger$ & 444    \\
    Avg. \# Words per Turn              & 60.63  \\
    Max. \# Words per Turn              & 474    \\
    Total \# Dialogues                  & 168    \\
    Total \# Turns                      & 1170   \\
    \bottomrule
  \end{tabular}
  \caption{Key statistics of \name.  Detailed statistics for individual tasks are provided in the Appendix. $\dagger$: Estimated using GPT-4 responses.}
  \label{tab:stat}
\end{table}

\section{\name}
\label{sec:benchmark}

{\name} is designed to comprehensively evaluate the multi-turn conversation capabilities of LLMs across a wide range of real-world application contexts.
By reviewing the existing authentic conversation datasets (like ShareGPT and LMSYS-Chat-1M dataset \citep{zheng_lmsys-chat-1m_2023}),
we have identified and categorized four primary modes of engagement in user-assistant interactions:

\textbf{Recollection}: Users present queries or tasks that necessitate the assistant's capacity to retrieve information from prior interactions, relying on the assistant's global context awareness and long-term memory capabilities. For instance, a user may instruct the model to initiate all the following responses with words starting with the letter "c."

\textbf{Expansion}: Users delve into various subjects while staying within the confines of the same topic.
For example, the user might ask different questions concerning one specific topic ``Michael Jordan''. Typically, the user will not refer to any specific details from previous dialogues.

\textbf{Refinement}: Users clarify or modify their previous instructions. For instance, users might add more detailed constraints, like specifying the desired output format, or provide feedback to clarify their instructions. This requires the assistant's ability to keep track of the instruction changes and leverage feedback to refine its responses.

\textbf{Follow-up}: Users ask questions that build upon the assistant's last response, often referencing specific details or opinions mentioned in that response. For instance, a user may seek additional information about a person mentioned in the assistant's prior response. This assesses the assistant's capacity to engage in coherent conversations.

  {\name} includes test sets targeting these four conversation categories (see Figure \ref{fig:four-task-examples}), while mirroring everyday scenarios of document processing, content creation, and information retrieval~\citep{zheng_lmsys-chat-1m_2023}.
It comprises 168 dialogue sessions with 1,170 turns to assess models' competence in handling such realistic multi-turn interactions.

\begin{table*}[!htbp]
  \vspace{-4mm}
  \centering
  \small
  \begin{tabular}{lccccc}
    \toprule
    Model                 & Avg.                & Recollection        & Expansion           & Refinement          & Follow-up           \\
    \midrule
    GPT-3.5-Turbo         & \highlighttwo{7.72} & 6.90                & \highlighttwo{7.87} & \highlighttwo{6.92} & 9.21                \\
    GPT-4                 & \highlight{9.03}    & \highlight{9.61}    & \highlight{9.07}    & \highlight{7.85}    & \highlight{9.60}    \\
    ChatGLM3-6B           & 5.49                & 2.92                & 5.90                & 4.73                & 8.39                \\
    Vicuna-7B-v1.5        & 6.44                & 5.45                & 6.70                & 5.31                & 8.31                \\
    Vicuna-13B-v1.5       & 7.01                & 6.27                & 6.70                & 6.37                & 8.68                \\
    Llama-2-chat-7B       & 6.11                & 3.86                & 5.87                & 6.20                & 8.53                \\
    Llama-2-chat-13B      & 6.31                & 3.66                & 6.37                & 6.37                & 8.82                \\
    Qwen-chat-7B          & 6.55                & 5.25                & 7.02                & 5.47                & 8.49                \\
    Qwen-chat-14B         & 7.26                & 6.21                & 7.58                & 6.11                & 9.12                \\
    Mistral-Instruct-7B   & 7.46                & \highlighttwo{7.22} & 6.98                & 6.58                & 9.05                \\
    Mixtral-Instruct-8x7B & 7.47                & 6.17                & 7.42                & 6.77                & \highlighttwo{9.52} \\
    \bottomrule
  \end{tabular}
  \caption{Multi-turn performance in four dialogue tasks. The highest score in each column is highlighted in \textbf{bold}, while the second-highest score is \uline{underlined}. Closed-source models outperform open-sourced models generally.
  }
  \label{tab:multi_turn_main_table}
  \vspace{-4mm}
\end{table*}
\subsection{Construction}
We extend existing datasets or construct new ones to prevent potential data leakage in \name.
When generating new instances, we use powerful LLMs like GPT-4 to produce initial instances.
Then, we manually evaluate each instance for difficulty, relevance, and originality, making revisions as necessary.

For Refinement and Expansion tasks, we convert several document-based NLP tasks into a dialogue format.
The underlying documents are also newly created with LLMs (an example prompt to instruct GPT-4 is detailed in Figure \ref{fig:construct_document}).
In the Refinement task, each dialogue turn introduces an additional constraint based on the previous instructions, whereas the Expansion task presents distinct tasks at each turn based on the same document introduced at the beginning.
We create two sub-tasks for the Recollection task with different difficulties. The first is a document classification task where the class labels are given in the first turn, while the second is a global instruction following task where the model must adhere to an instruction given in the first turn throughout the dialogue \citep{zhou_instruction-following_2023}.
Regarding the Follow-up task, we extend MT-Bench \citep{zheng_judging_2023} by adding three extra turns, each of which poses a question based on the previous assistant's response.

We also establish the corresponding single-turn instances for all separate turns in all tasks except Follow-up, which inherently depends on the assistant's previous response and difficult to construct the equivalent single-turn instances.
To construct the single-turn instances in the Expansion and Refinement task, we extract each turn together with the relevant context (like the relevant document and instruction) into one single turn.
For the Refinement task, we utilize GPT-4 to condense the multiple instructions into one.
Specific details for constructing each task can be found in Appendix \ref{sec:benchmark_detail}.

\section{Experiment}

\begin{table*}[!htbp]
  \centering
  \small
  \begin{tabular}{lcrcccccc}
    \toprule
    \multirow{2}{*}{Model} & \multirow{2}{*}{\thead{ST  Avg.}} & \multirow{2}{*}{\thead{MT Avg.}} & \multicolumn{2}{c}{Recollection} & \multicolumn{2}{c}{Expansion} & \multicolumn{2}{c}{Refinement}                                                                   \\
    \cmidrule(lr){4-5} \cmidrule(lr){6-7} \cmidrule(lr){8-9}
                           &                                   &                                  & ST                               & MT                            & ST                             & MT                  & ST                  & MT                  \\
    \midrule
    GPT-3.5-Turbo          & 8.07                              & \highlighttwo{7.23} (-0.84)      & \highlighttwo{8.75}              & 6.90                          & 8.39                           & \highlighttwo{7.87} & 7.08                & \highlighttwo{6.92} \\
    GPT-4                  & \highlight{9.17}                  & \highlight{8.84} (-0.33)         & \highlight{9.53}                 & \highlight{9.61}              & \highlight{9.53}               & \highlight{9.07}    & \highlight{8.46}    & \highlight{7.85}    \\
    ChatGLM3-6B            & 5.71                              & 4.52 (-1.19)                     & 5.05                             & 2.92                          & 7.20                           & 5.90                & 4.89                & 4.73                \\
    Vicuna-7B-v1.5         & 6.31                              & 5.82 (-0.49)                     & 6.35                             & 5.45                          & 6.99                           & 6.70                & 5.60                & 5.31                \\
    Vicuna-13B-v1.5        & 7.10                              & 6.45 (-0.65)                     & 6.98                             & 6.27                          & 7.67                           & 6.70                & 6.66                & 6.37                \\
    Llama-2-chat-7B        & 7.21                              & 5.31 (-1.90)                     & 7.26                             & 3.86                          & 7.36                           & 5.87                & 7.00                & 6.20                \\
    Llama-2-chat-13B       & 7.55                              & 5.47 (-2.08)                     & 7.51                             & 3.66                          & 7.86                           & 6.37                & 7.29                & 6.37                \\
    Qwen-chat-7B           & 6.86                              & 5.91 (-0.95)                     & 7.17                             & 5.25                          & 7.46                           & 7.02                & 5.96                & 5.47                \\
    Qwen-chat-14B          & 7.62                              & 6.64 (-0.98)                     & 8.40                             & 6.21                          & 7.90                           & 7.58                & 6.58                & 6.11                \\
    Mistral-Instruct-7B    & 7.69                              & 6.93 (-0.76)                     & 8.47                             & \highlighttwo{7.22}           & 7.60                           & 6.98                & 7.00                & 6.58                \\
    Mixtral-Instruct-8x7B  & \highlighttwo{8.28}               & 6.78 (-1.50)                     & 7.86                             & 6.17                          & \highlighttwo{9.50}            & 7.42                & \highlighttwo{7.48} & 6.77                \\
    \bottomrule
  \end{tabular}
  \caption{Performance of various models across different dialogue tasks in both single-turn and multi-turn settings. \textbf{ST} and \textbf{MT} denote single-turn and multi-turn respectively. The best score in each column is highlighted in \textbf{bold} and the second-highest score is \uline{underlined}. Bracketed numbers indicate the change in score between the single-turn and multi-turn scenarios.
    The Follow-up task is omitted since there is no equivalent single-turn setting.
    Most models exhibit a substantial performance gap between the single-turn and multi-turn settings.
  }
  \label{tab:main_table}
  \vspace{-5mm}

\end{table*}

\subsection{Models}
We conduct evaluations on 11 popular LLMs, including ChatGLM3-6B \citep{du_glm_2022}, Vicuna-v1.5 (7B, 13B) \citep{vicuna2023}, Llama-2-chat (7B, 13B) \citep{touvron_llama_2023}, Qwen-chat (7B, 14B) \citep{bai_qwen_2023}, Mistral-Instruct-7B \citep{jiang_mistral_2023}, Mixtral-Instruct-8x7B \citep{jiang_mixtral_2024}, GPT-3.5-Turbo \citep{ouyang_training_2022}, and GPT-4 \citep{openai_gpt-4_2023}\footnote{We utilized gpt-3.5-turbo-0613 and gpt-4-0613 from \url{https://cuhk-api-dev1-apim1.developer.azure-api.net}.}.

\subsection{Implementation Details}
We use the corresponding chatting format for each LLM in all experiments\footnote{We used FastChat \url{https://github.com/lm-sys/FastChat}, which provides the prompt format for various LLMs.}.
To ensure reproducibility, we employ greedy decoding for both inference and evaluation.
During inference, we use the system prompt "You are a helpful, respectful and honest assistant." for all models, while an empty system prompt is used for evaluation.

\subsection{Evaluation}
Evaluating LLM responses poses challenges as the additional content is often included in the generated responses, such as introductions, conclusions, or supplementary explanations \citep{yue_mmmu_2023,zhou_instruction-following_2023}. This hinders accurate quality assessment using rule-based automatic evaluation metrics.
Recent research shows that LLM-based evaluation, especially utilizing GPT-4, achieves high agreement with human judgment \citep{zheng_judging_2023,bitton_visit-bench_2023}.
In light of this, we utilize GPT-4 to evaluate all responses, except for classification and recollection tasks, which can be scored with simple rules.

To evaluate the responses, we use a zero-shot prompt to guide GPT-4 to assign an integer rating from 1 to 10 based on the relevant context.
We ask GPT-4 to provide a rationale before the score, following a chain-of-thought approach \citep{wei_chain--thought_2022}.
Previous work has also shown that utilizing chain-of-thought in evaluation enhances the quality \citep{liu_g-eval_2023}.
The evaluation prompt can be found in Figure \ref{fig:prompt_evaluation}.

We evaluate the global following sub-task in the Recollection task using heuristics and rules \citep{zhou_instruction-following_2023}, calculating the average number of dialogue turns adhering to the global instruction and normalizing the result to a maximum score of 10.
For the document classification task, we measure the classification accuracy directly and normalize it to a full score of 10.

A good multi-turn conversational model should demonstrate strong capacity in multi-turn interactions and exhibit a minimal performance gap between single-turn and multi-turn settings.
Therefore, we also evaluate the corresponding single-turn performance, in addition to multi-turn settings, to measure the gap between them.

\subsection{Result}

Table \ref{tab:multi_turn_main_table} shows the multi-turn performance of the evaluated LLMs across all four task categories. With the exception of GPT-4, all models achieve an average score lower than 8, indicating that {\name} poses a considerable challenge for multi-turn capabilities.
In particular, most models perform worst in the Recollection task, failing to obey the global instruction stated initially in successive turns.
All models also perform poorly in the Refinement task, often ignoring constraints from previous turns.
Overall, {\name} comprises tasks of varying difficulty, targeting various aspects of multi-turn interaction and effectively highlighting the strengths and weaknesses of LLMs across diverse multi-turn scenarios. We provide a more detailed discussion of the results below.

\paragraph{Performance of Closed- versus Open-Source LLMs.}
Table \ref{tab:multi_turn_main_table} illustrates that closed-source LLMs, including GPT-3.5-Turbo and GPT-4, generally outperform open-source ones in multi-turn dialogues.
GPT-4 stands out with an impressive average score of 9.03, significantly surpassing the second-best model's score of 7.72.
However, it's noteworthy that open-source models like Mistral-Instruct-7B and Mixtral-Instruct-8x7B showcase exceptional performance in specific tasks, putting them on par with or even surpassing GPT-3.5-Turbo's performance. For instance, Mixtral-Instruct-8x7B achieves a score of 9.52 in Follow-up dialogues, outperforming GPT-3.5-Turbo's score of 9.21. These findings align with recent research, which suggests that open-source LLMs can achieve comparable or even superior performance to closed-source LLMs in certain domains \citep{chen_chatgpts_2023}.
Mistral-Instruct-7B surpasses all 7B models and exhibits performance comparable to 13B models, consistent with the findings reported in \citet{jiang_mistral_2023}.
On the other hand, Mixtral-Instruct-8x7B, despite its strong performance in most tasks, faces challenges in adhering to global instructions in recollection tasks, resulting in a similar average score as Mistral-Instruct-7B.

\paragraph{Inferior Performance in Multi-Turn Dialogues.}
Table \ref{tab:main_table} presents the evaluation results of both multi-turn and single-turn instances, highlighting the performance gap between them.
The results indicate that most models exhibit a substantial decline in the performance of multi-turn dialogues compared to single-turn instances.
This performance gap therefore serves as a valuable indicator of a model's multi-turn capabilities.
Notably, the observed gap between the two scenarios does not appear to be directly correlated with the fundamental capabilities of the models.
For instance, while Llama-2-chat models outperform Vicuna models in the single-turn setting, they noticeably lag behind in multi-turn dialogues.
This observation underscores the importance of including multi-turn evaluation when conducting a comprehensive evaluation of LLMs.

\paragraph{Challenges in Long-Distance Information Retrieval for LLMs.}
Our study reveals that LLMs often underperform in tasks requiring information from earlier dialogue turns.
In the Recollection task, all LLMs except GPT-4 struggle to adhere to the initial global instructions as the conversation length, i.e., distance from their initial instruction, increases.
Table \ref{tab:refinement} also supports this trend, revealing that most models perform better on the first task (i.e., the first six turns) compared to the second (i.e., the final six turns), as the turns in the second task are further from the given document at the beginning.
Our error analysis confirms that LLMs commonly overlook prior instructions. A detailed discussion is presented in Section \ref{sec:further_analysis}.

\begin{figure}[ht]
  \centering
  \includegraphics[width=\columnwidth]{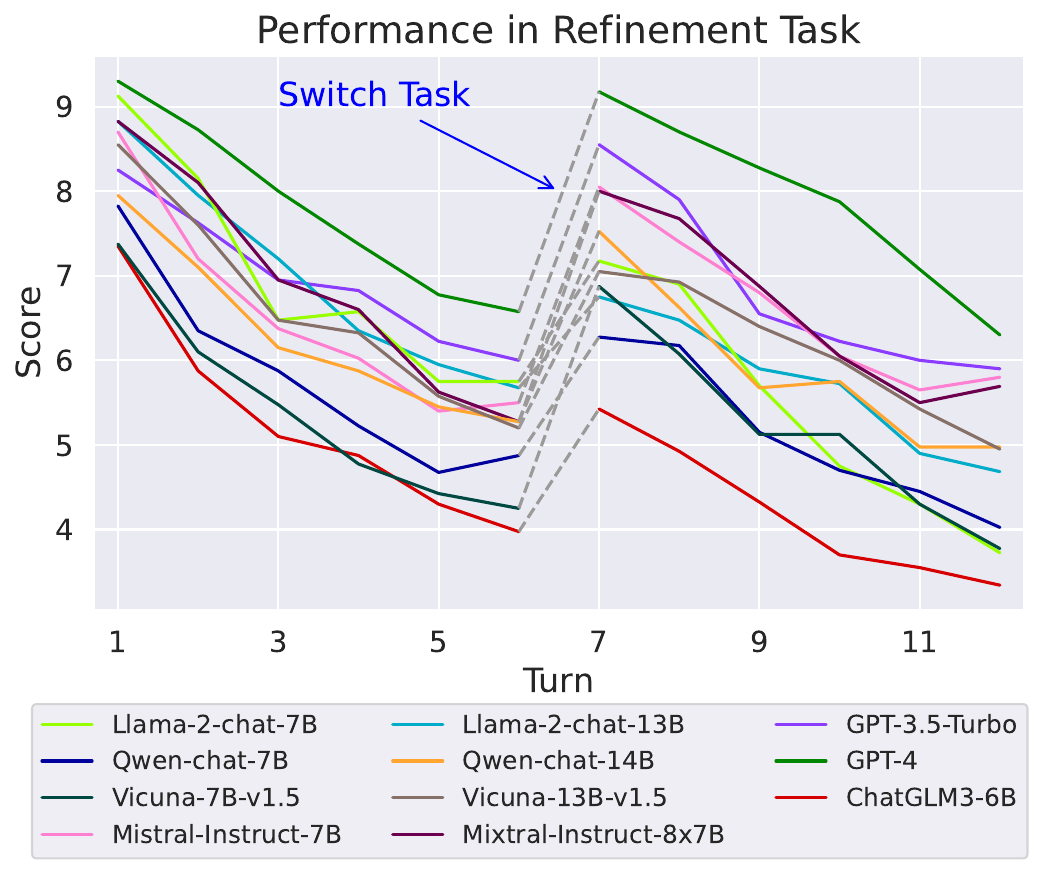}
  \caption{Performance across turns in Refinement task. Each dialogue has two NLP tasks with each task comprising six increasingly complex instructions. The transition to the second NLP task occurs at the seventh turn as denoted by the grey dashed line. The performance of all models declines as more instructions are added. }

  \label{fig:refinement_turn}
\end{figure}

\begin{table}[ht]
  \centering
  \small
  \begin{tabular}{lccc}
    \toprule
                          & First & Second & Difference \\
    \midrule
    GPT-3.5-Turbo         & 6.98  & 6.85   & -0.12      \\
    GPT-4                 & 7.79  & 7.90   & 0.11       \\
    ChatGLM3-6B           & 5.25  & 4.21   & -1.03      \\
    Vicuna-7B-v1.5        & 5.40  & 5.21   & -0.19      \\
    Vicuna-13B-v1.5       & 6.62  & 6.12   & -0.50      \\
    Llama-2-chat-7B       & 6.97  & 5.42   & -1.55      \\
    Llama-2-chat-13B      & 6.99  & 5.74   & -1.25      \\
    Qwen-chat-7B          & 5.80  & 5.13   & -0.67      \\
    Qwen-chat-14B         & 6.30  & 5.92   & -0.38      \\
    Mistral-Instruct-7B   & 6.53  & 6.62   & 0.09       \\
    Mixtral-Instruct-8x7B & 6.90  & 6.63   & -0.26      \\
    \bottomrule
  \end{tabular}
  \caption{Performance of the first task (the first six turns) and the second task (last six turns) in Refinement task. The performance difference between the two tasks is also shown.}
  \vspace{-5mm}
  \label{tab:refinement}
\end{table}

\subsection{Further Analysis}
\label{sec:further_analysis}
This section presents further analyses of five top-performing models.
We meticulously examine the ten responses with the largest score difference in multi-turn and single-turn scenarios generated by each model across the four dialogue tasks, resulting in an analysis of a total of 200 responses.
The objective is to understand the factors that degrade model performance in multi-turn dialogues, rather than assessing their foundational capabilities.
Our analysis reveals that 99 responses (49.5\%) did not comply with earlier instructions,  96 responses (48\%) were misdirected by the errors accumulated in the earlier context, and 5 instances (2.5\%) were attributed to evaluation errors. A comprehensive analysis of these findings is provided below.

\begin{figure}[htbp]
  \centering
  \vspace{-10mm}
  \includegraphics[width=.9\columnwidth]{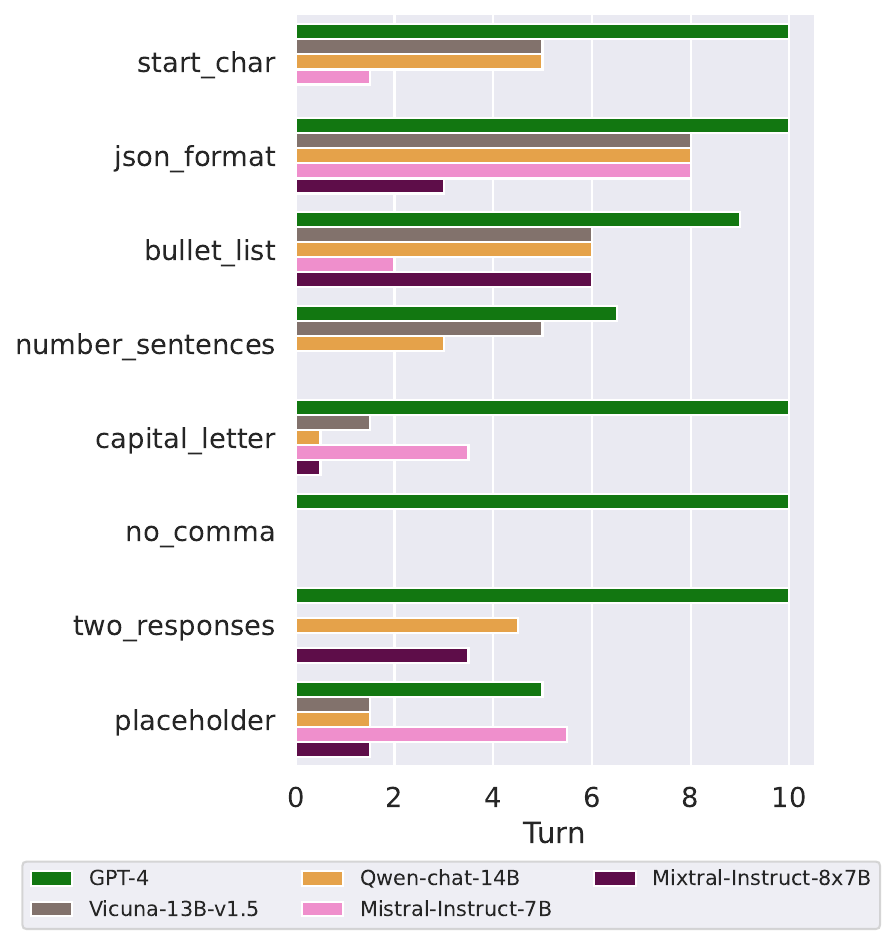}
  \vspace{-2mm}
  \caption{The average number of turns that different models can adhere to the instructions in the Recollection task. Each instruction consists of two dialogue sessions with ten dialogue turns. The description of the instructions can be found in Table \ref{tab:turns_in_recollection} in the Appendix.}
  \label{fig:turns_in_recollection}
\end{figure}

\begin{table}[!htbp]
  \centering
  \small
  \vspace{-1mm}
  \begin{tabular}{lcc}
    \toprule
    Task       & Pearson & Spearman \\
    \midrule
    Refinement & 0.68    & 0.52     \\
    Expansion  & 0.61    & 0.59     \\
    Follow-up  & 0.66    & 0.64     \\
    \midrule
    Avg.       & 0.65    & 0.58     \\
    \bottomrule
  \end{tabular}
  \caption{The correlation scores between human ratings and GPT-4 ratings for different tasks.}
  \label{tab:rating_correlation}
  \vspace{-5mm}
\end{table}

\begin{table*}[!htbp]
  \centering
  \small
  \begin{tabular}{lc|ccc|ccc}
    \toprule
                          & Without          & 1 Between        & 3 Between        & 6 Between        & 1 Front          & 3 Front          & 6 Front          \\
    \midrule
    GPT-4                 & \highlight{7.79} & \highlight{7.90} & \highlight{7.76} & \highlight{7.97} & \highlight{7.65} & \highlight{7.60} & \highlight{7.65} \\
    Vicuna-13B-v1.5       & 6.62             & 5.91             & 5.47             & 5.56             & 6.25             & 6.16             & 5.89             \\
    Qwen-chat-14B         & 6.30             & 5.89             & 5.76             & 5.17             & 6.22             & 6.01             & 6.18             \\
    Mistral-Instruct-7B   & 6.53             & 6.44             & 6.25             & 6.08             & 6.66             & 6.68             & 6.83             \\
    Mixtral-Instruct-8x7B & 6.90             & 6.47             & 6.57             & 6.33             & 7.01             & 6.58             & 6.89             \\
    \bottomrule
  \end{tabular}
  \vspace{-2mm}
  \caption{Performance of various LLMs in Refinement task with varying numbers of distracting turns (1, 3, or 6) inserted at the front (Front) or in between (Between) the document and query turns. }
  \label{tab:ablation_irrelevant}
\end{table*}

\begin{table*}[!htbp]
  \centering
  \small
  \begin{tabular}{lcccccc}
    \toprule
    \multirow{2}{*}{Model} & \multicolumn{2}{c}{Recollection} & \multicolumn{2}{c}{Expansion} & \multicolumn{2}{c}{Refinement}                                                          \\
    \cmidrule(lr){2-3} \cmidrule(lr){4-5} \cmidrule(lr){6-7}
                           & Predicted                        & Gold                          & Predicted                      & Gold             & Predicted        & Gold             \\
    \midrule
    GPT-4                  & \highlight{9.71}                 & \highlight{9.89}              & \highlight{9.07}               & \highlight{9.20} & \highlight{7.85} & \highlight{7.86} \\
    Vicuna-13B-v1.5        & 4.64                             & 7.32                          & 6.70                           & 6.87             & 6.37             & 7.15             \\
    Qwen-chat-14B          & 4.43                             & 7.00                          & 7.58                           & 7.63             & 6.11             & 6.95             \\
    Mistral-Instruct-7B    & 5.25                             & 7.29                          & 6.98                           & 7.02             & 6.58             & 7.38             \\
    Mixtral-Instruct-8x7B  & 3.21                             & 7.11                          & 7.42                           & 7.47             & 6.77             & 7.17             \\
    \bottomrule
  \end{tabular}
  \vspace{-2mm}
  \caption{Comparison of model performance in three dialogue tasks, conditioned on dialogue history with self-generated responses versus gold responses.}
  \label{tab:ablation_gold}
  \vspace{-2mm}
\end{table*}

\paragraph{Noncompliance with Earlier Instructions (49.5\%).}
Figure \ref{fig:turns_in_recollection} shows a significant variation in how models follow the initial instructions in the Recollection task.
While all open-source models encounter challenges with specific instructions, GPT-4 consistently adheres to the majority, especially those prohibiting comma usage or requiring generating two distinct responses simultaneously.
Mixtral-Instruct-8x7B, having a strong performance in other multi-turn tasks, struggles with following many global instructions like formatting responses as JSON. Our case studies also reveal that LLMs often forget previous instructions. An example is illustrated in Figure \ref{fig:case_refine}.

All models encounter difficulties with counting-related instructions, like limiting responses to a specific number of sentences or including a set number of placeholders in the response. This issue also arises in the Refinement task, where models often struggle to identify the correct paragraph for tasks such as translation or noun identification in the n-th paragraph.

\paragraph{Error Propagation (48\%).}
Accumulated errors from preceding dialogue turns often confuse the models, leading to more incorrect responses.
A notable example of this issue is the misidentification of the correct paragraph in the initial turn of many refinement tasks.
The models persist in fulfilling new instructions based on this incorrect paragraph, which further accumulate errors and result in consistently low scores throughout the dialogue.
We explore this phenomenon in greater depth through ablation studies detailed in Section \ref{sec:ablation}.

\paragraph{Evaluation (2.5\%).}
GPT-4 occasionally misinterprets instructions and provides inaccurate evaluations, particularly in the Refinement task when presented with a series of instructions.
Figure \ref{fig:eval_wrong} presents an example where GPT-4 fails to identify the relevant constraints, despite being prompted to disregard irrelevant constraints from earlier turns.

To verify that GPT-4's evaluation mostly aligns with human preference, we recruited 5 annotators to evaluate 60 randomly selected responses from each multi-turn dialogue task, excluding the Recollection task which uses automatic evaluations. They evaluated in total of 180 responses. More details regarding the human evaluation can be found in Appendix \ref{sec:human_evaluation}.

Table \ref{tab:rating_correlation} shows Pearson's correlation and Spearman's rank correlation coefficient between human ratings and GPT-4 ratings. The average Spearman correlation of 0.58 exceeds the scores of other evaluation methods for various natural language generation tasks \citep{liu_g-eval_2023}.
The results indicate that GPT-4 ratings align well with human ratings, consistent with recent findings \citep{zheng_judging_2023,bitton_visit-bench_2023}.

\subsection{Ablation Study}
\label{sec:ablation}
\begin{table*}[!htbp]
  \centering
  \small
  \begin{tabular}{lcccccc|c}
    \toprule
                          & Gold              & DCG               & SCG               & RC                & RC (5)            & RC (10)           & ST                \\
    \midrule
    GPT-4                 & \highlight{98.00} & \highlight{98.00} & \highlight{97.00} & \highlight{89.00} & \highlight{96.00} & \highlight{82.00} & 93.00             \\
    Vicuna-13B-v1.5       & 81.00             & 84.00             & 70.00             & 45.00             & 62.00             & 28.00             & 75.00             \\
    Qwen-chat-14B         & 94.00             & 95.00             & 86.00             & 69.00             & 68.00             & 60.00             & \highlight{94.00} \\
    Mistral-Instruct-7B   & 96.00             & 95.00             & 95.00             & 75.00             & 80.00             & 70.00             & \highlight{94.00} \\
    Mixtral-Instruct-8x7B & 95.00             & 95.00             & 94.00             & 57.00             & 60.00             & 54.00             & 88.00             \\
    \bottomrule
  \end{tabular}
  \vspace{-2mm}
  \caption{Performance in classification task using various dialogue contexts.
    \textit{Gold}: Randomly select documents with their proper labels.
    \textit{DCG}: The Diverse Class Gold setting. Similar to \textit{Gold}, but exclude documents with the same labels as the current turn.
    \textit{SCG}: The Single Class Gold setting, which randomly chooses documents from the same category, avoiding the current turn's label.
    \textit{RC}: The Random Class setting, which randomly selects documents and assigns random labels to them. The \textit{(5)} and \textit{(10)} refer to the performance at turn 5 and 10 respectively.
    \textit{ST}: Single-Turn, the single-turn performance with no dialogue context.}
  \label{tab:ablation_cls}
  \vspace{-4mm}

\end{table*}
Inspired by the insights gained from the earlier sections, we proceed to conduct two ablations studies to investigate the effects of varying dialogue contexts on model's performance.
Additionally, we conduct another ablation study to explore how the distance between the relevant context and the current query affects performance.

\paragraph{Gold Context vs. Self-Predicted Context.}
Table \ref{tab:ablation_gold} presents the results of three dialogue tasks, conditioned on dialogue history of self-generated responses (i.e. the main results) or gold responses from human verified GPT-4 outputs.
The results indicate that models conditioned on gold context exhibit significant improvement in Recollection and Refinement tasks.
We attribute this performance gap to two factors.
Firstly, using gold context prevents the error propagation from earlier turns. This can be confirmed by the minor gap observed in GPT-4, the most powerful model, which makes much fewer errors.
Secondly, the gold responses serve as in-context examples, providing valuable knowledge for the model \citep{brown_language_2020}.
Notably, using gold responses in the Expansion task yields only a slight improvement.
This is likely because each dialogue turn in this task is a distinct NLP task, thus not benefiting from these examples of other tasks.

\paragraph{Influence of Dialogue History as In-context Examples.}
We investigate the impact of dialogue history as in-context examples on model performance in document classification, following previous work \citep{minRethinkingRoleDemonstrations2022}.
We manipulate dialogue history in four settings and vary the number of dialogue turns (either four or nine).
Each turn includes a document, and a category depending on the setting used.
The \textit{Gold} setting involves random documents with their correct labels.
The \textit{Diverse Class Gold} setting is similar to \textit{Gold} but excludes documents sharing the current turn's label.
In the \textit{Single Class Gold} setting, documents from a randomly chosen category are provided, avoiding the current turn's label.
The \textit{Random Class} setting assigns random labels to the randomly selected documents.
\textit{Random Class (5)} and \textit{Random Class (10)} denotes the performance in turn 5 and 10 respectively.

Our findings in Table \ref{tab:ablation_cls} reveal that incorporating gold labels of randomly selected documents (\textit{Gold}) improves performance compared to zero-shot setting, underscoring the value of in-context examples in dialogue history \citep{brown_language_2020,minRethinkingRoleDemonstrations2022}.
The \textit{Diverse Class Gold} setting yields similar improvements, even with the documents belonging to the same label as the current turn excluded in the dialogue history.
However, dialogue history limited to a single class can negatively impact weaker models, suggesting that biased examples may be harmful.
Contrary to previous findings \citep{minRethinkingRoleDemonstrations2022}, the \textit{Random Class} setting significantly reduces performance.
Furthermore, the performance at the 10th turn is even worse than the 5th turn, indicating the presence of error propagation.
We extend our analysis to the Follow-up task with a more complicated setup. The detailed analysis can be found in Appendix \ref{sec:ablation_followup}.

\paragraph{Impact of Irrelevant Context on Performance.}
We conduct further experiments to examine how irrelevant context, placed at different positions, affects the performance in multi-turn dialogues.
We insert varying numbers of dialogue turns, randomly sampled from LMSYS-Chat-1M \citep{zheng_lmsys-chat-1m_2023}, either at the beginning or between the document and the query turns in the Refinement task.

Table \ref{tab:ablation_irrelevant} shows that inserting these turns at the beginning results in mixed outcomes.
Notably, Mistral-Instruct-7B and Mixtral-Instruct-8x7B even show improved performance, while other models show slight declines.
This suggests that models are capable of switching topics in a multi-turn dialogue without being affected by previous discussions.
Conversely, inserting distracting turns between the document and query turns consistently degrades performance (GPT-4 is an exception, as it is minimally affected by any circumstances, demonstrating its strong resistance to interference).
This further supports that the increasing distance between the document and the queries negatively impacts performance in multi-turn dialogues.

\section{Conclusion}
 {\name} represents an important first step in systematically evaluating and understanding LLMs' multi-turn conversational abilities.
Our experiment shows a pronounced gap between single-turn versus multi-turn performance across current models, a phenomenon that persists irrespective of the underlying capabilities of the models.
Our comprehensive analysis reveals that the distance to relevant content and susceptibility to error propagation are the key factors that cause a decline in multi-turn performance.
We believe this work not only sheds light on the current limitations of LLM's multi-turn conversational abilities, it also paves the way for further efforts to close the identified gap and develop robust conversational models capable of multi-turn interactions.

\section*{Limitations}
While GPT-4's evaluation has been shown to align well with human ratings, evaluation errors persist.
This work employs chain-of-thought in evaluation \citep{wei_chain--thought_2022}, which enhances quality.
To further improve the evaluation quality, more advanced prompting or reasoning techniques can be explored, like tree-of-thought \citep{yao_tree_2023} or self-consistency \citep{chen_universal_2023}. While these methods require more model inference times, this is a trade-off between evaluation quality and cost.

Due to computational limits, our experiments did not include any larger open-source models like Llama2-chat-70B. Further studies can investigate whether larger LLMs exhibit similar findings outlined in this paper.

\bibliography{anthology,custom,zotero}

\begin{thebibliography}{26}
\expandafter\ifx\csname natexlab\endcsname\relax\def\natexlab#1{#1}\fi

\bibitem[{Bai et~al.(2023)Bai, Bai, Chu, Cui, Dang, Deng, Fan, Ge, Han, Huang, Hui, Ji, Li, Lin, Lin, Liu, Liu, Lu, Lu, Ma, Men, Ren, Ren, Tan, Tan, Tu, Wang, Wang, Wang, Wu, Xu, Xu, Yang, Yang, Yang, Yang, Yao, Yu, Yuan, Yuan, Zhang, Zhang, Zhang, Zhang, Zhou, Zhou, Zhou, and Zhu}]{bai_qwen_2023}
Jinze Bai, Shuai Bai, Yunfei Chu, Zeyu Cui, Kai Dang, Xiaodong Deng, Yang Fan, Wenbin Ge, Yu~Han, Fei Huang, Binyuan Hui, Luo Ji, Mei Li, Junyang Lin, Runji Lin, Dayiheng Liu, Gao Liu, Chengqiang Lu, Keming Lu, Jianxin Ma, Rui Men, Xingzhang Ren, Xuancheng Ren, Chuanqi Tan, Sinan Tan, Jianhong Tu, Peng Wang, Shijie Wang, Wei Wang, Shengguang Wu, Benfeng Xu, Jin Xu, An~Yang, Hao Yang, Jian Yang, Shusheng Yang, Yang Yao, Bowen Yu, Hongyi Yuan, Zheng Yuan, Jianwei Zhang, Xingxuan Zhang, Yichang Zhang, Zhenru Zhang, Chang Zhou, Jingren Zhou, Xiaohuan Zhou, and Tianhang Zhu. 2023.
\newblock \href {https://doi.org/10.48550/arXiv.2309.16609} {Qwen {{Technical Report}}}.

\bibitem[{Bitton et~al.(2023)Bitton, Bansal, Hessel, Shao, Zhu, Awadalla, Gardner, Taori, and Schmidt}]{bitton_visit-bench_2023}
Yonatan Bitton, Hritik Bansal, Jack Hessel, Rulin Shao, Wanrong Zhu, Anas Awadalla, Josh Gardner, Rohan Taori, and Ludwig Schmidt. 2023.
\newblock \href {https://doi.org/10.48550/arXiv.2308.06595} {{{VisIT-Bench}}: {{A Benchmark}} for {{Vision-Language Instruction Following Inspired}} by {{Real-World Use}}}.

\bibitem[{Brown et~al.(2020)Brown, Mann, Ryder, Subbiah, Kaplan, Dhariwal, Neelakantan, Shyam, Sastry, Askell, Agarwal, {Herbert-Voss}, Krueger, Henighan, Child, Ramesh, Ziegler, Wu, Winter, Hesse, Chen, Sigler, Litwin, Gray, Chess, Clark, Berner, McCandlish, Radford, Sutskever, and Amodei}]{brown_language_2020}
Tom Brown, Benjamin Mann, Nick Ryder, Melanie Subbiah, Jared~D Kaplan, Prafulla Dhariwal, Arvind Neelakantan, Pranav Shyam, Girish Sastry, Amanda Askell, Sandhini Agarwal, Ariel {Herbert-Voss}, Gretchen Krueger, Tom Henighan, Rewon Child, Aditya Ramesh, Daniel Ziegler, Jeffrey Wu, Clemens Winter, Chris Hesse, Mark Chen, Eric Sigler, Mateusz Litwin, Scott Gray, Benjamin Chess, Jack Clark, Christopher Berner, Sam McCandlish, Alec Radford, Ilya Sutskever, and Dario Amodei. 2020.
\newblock \href {https://proceedings.neurips.cc/paper/2020/hash/1457c0d6bfcb4967418bfb8ac142f64a-Abstract.html} {Language {{Models}} are {{Few-Shot Learners}}}.
\newblock In \emph{Advances in {{Neural Information Processing Systems}}}, volume~33, pages 1877--1901. {Curran Associates, Inc.}

\bibitem[{Chen et~al.(2023{\natexlab{a}})Chen, Jiao, Li, Qin, Ravaut, Zhao, Xiong, and Joty}]{chen_chatgpts_2023}
Hailin Chen, Fangkai Jiao, Xingxuan Li, Chengwei Qin, Mathieu Ravaut, Ruochen Zhao, Caiming Xiong, and Shafiq Joty. 2023{\natexlab{a}}.
\newblock \href {https://arxiv.org/abs/2311.16989v3} {{{ChatGPT}}'s {{One-year Anniversary}}: {{Are Open-Source Large Language Models Catching}} up?}

\bibitem[{Chen et~al.(2023{\natexlab{b}})Chen, Aksitov, Alon, Ren, Xiao, Yin, Prakash, Sutton, Wang, and Zhou}]{chen_universal_2023}
Xinyun Chen, Renat Aksitov, Uri Alon, Jie Ren, Kefan Xiao, Pengcheng Yin, Sushant Prakash, Charles Sutton, Xuezhi Wang, and Denny Zhou. 2023{\natexlab{b}}.
\newblock \href {https://doi.org/10.48550/arXiv.2311.17311} {Universal {{Self-Consistency}} for {{Large Language Model Generation}}}.

\bibitem[{Chiang et~al.(2023)Chiang, Li, Lin, Sheng, Wu, Zhang, Zheng, Zhuang, Zhuang, Gonzalez, Stoica, and Xing}]{vicuna2023}
Wei-Lin Chiang, Zhuohan Li, Zi~Lin, Ying Sheng, Zhanghao Wu, Hao Zhang, Lianmin Zheng, Siyuan Zhuang, Yonghao Zhuang, Joseph~E. Gonzalez, Ion Stoica, and Eric~P. Xing. 2023.
\newblock \href {https://lmsys.org/blog/2023-03-30-vicuna/} {Vicuna: An open-source chatbot impressing gpt-4 with 90\%* chatgpt quality}.

\bibitem[{Du et~al.(2022)Du, Qian, Liu, Ding, Qiu, Yang, and Tang}]{du_glm_2022}
Zhengxiao Du, Yujie Qian, Xiao Liu, Ming Ding, Jiezhong Qiu, Zhilin Yang, and Jie Tang. 2022.
\newblock \href {https://doi.org/10.18653/v1/2022.acl-long.26} {{{GLM}}: {{General Language Model Pretraining}} with {{Autoregressive Blank Infilling}}}.
\newblock In \emph{Proceedings of the 60th {{Annual Meeting}} of the {{Association}} for {{Computational Linguistics}} ({{Volume}} 1: {{Long Papers}})}, pages 320--335, {Dublin, Ireland}. {Association for Computational Linguistics}.

\bibitem[{Hendrycks et~al.(2020)Hendrycks, Burns, Basart, Zou, Mazeika, Song, and Steinhardt}]{hendrycks_measuring_2020}
Dan Hendrycks, Collin Burns, Steven Basart, Andy Zou, Mantas Mazeika, Dawn Song, and Jacob Steinhardt. 2020.
\newblock \href {https://openreview.net/forum?id=d7KBjmI3GmQ} {Measuring {{Massive Multitask Language Understanding}}}.
\newblock In \emph{International {{Conference}} on {{Learning Representations}}}.

\bibitem[{Jiang et~al.(2023)Jiang, Sablayrolles, Mensch, Bamford, Chaplot, de~las Casas, Bressand, Lengyel, Lample, Saulnier, Lavaud, Lachaux, Stock, Scao, Lavril, Wang, Lacroix, and Sayed}]{jiang_mistral_2023}
Albert~Q. Jiang, Alexandre Sablayrolles, Arthur Mensch, Chris Bamford, Devendra~Singh Chaplot, Diego de~las Casas, Florian Bressand, Gianna Lengyel, Guillaume Lample, Lucile Saulnier, L{\'e}lio~Renard Lavaud, Marie-Anne Lachaux, Pierre Stock, Teven~Le Scao, Thibaut Lavril, Thomas Wang, Timoth{\'e}e Lacroix, and William~El Sayed. 2023.
\newblock \href {https://doi.org/10.48550/arXiv.2310.06825} {Mistral {{7B}}}.

\bibitem[{Jiang et~al.(2024)Jiang, Sablayrolles, Roux, Mensch, Savary, Bamford, Chaplot, de~las Casas, Hanna, Bressand, Lengyel, Bour, Lample, Lavaud, Saulnier, Lachaux, Stock, Subramanian, Yang, Antoniak, Scao, Gervet, Lavril, Wang, Lacroix, and Sayed}]{jiang_mixtral_2024}
Albert~Q. Jiang, Alexandre Sablayrolles, Antoine Roux, Arthur Mensch, Blanche Savary, Chris Bamford, Devendra~Singh Chaplot, Diego de~las Casas, Emma~Bou Hanna, Florian Bressand, Gianna Lengyel, Guillaume Bour, Guillaume Lample, L{\'e}lio~Renard Lavaud, Lucile Saulnier, Marie-Anne Lachaux, Pierre Stock, Sandeep Subramanian, Sophia Yang, Szymon Antoniak, Teven~Le Scao, Th{\'e}ophile Gervet, Thibaut Lavril, Thomas Wang, Timoth{\'e}e Lacroix, and William~El Sayed. 2024.
\newblock \href {https://doi.org/10.48550/arXiv.2401.04088} {Mixtral of {{Experts}}}.

\bibitem[{Lee et~al.(2023)Lee, Srivastava, Hardy, Thickstun, Durmus, Paranjape, {Gerard-Ursin}, Li, Ladhak, Rong, Wang, Kwon, Park, Cao, Lee, Bommasani, Bernstein, and Liang}]{lee_evaluating_2023}
Mina Lee, Megha Srivastava, Amelia Hardy, John Thickstun, Esin Durmus, Ashwin Paranjape, Ines {Gerard-Ursin}, Xiang~Lisa Li, Faisal Ladhak, Frieda Rong, Rose~E. Wang, Minae Kwon, Joon~Sung Park, Hancheng Cao, Tony Lee, Rishi Bommasani, Michael~S. Bernstein, and Percy Liang. 2023.
\newblock \href {https://openreview.net/forum?id=hjDYJUn9l1} {Evaluating {{Human-Language Model Interaction}}}.
\newblock \emph{Transactions on Machine Learning Research}.

\bibitem[{Liao et~al.(2023)Liao, Meng, Liu, Wang, and Wang}]{liao2023automatic}
Yusheng Liao, Yutong Meng, Hongcheng Liu, Yanfeng Wang, and Yu~Wang. 2023.
\newblock An automatic evaluation framework for multi-turn medical consultations capabilities of large language models.
\newblock \emph{arXiv preprint arXiv:2309.02077}.

\bibitem[{Liu et~al.(2023)Liu, Iter, Xu, Wang, Xu, and Zhu}]{liu_g-eval_2023}
Yang Liu, Dan Iter, Yichong Xu, Shuohang Wang, Ruochen Xu, and Chenguang Zhu. 2023.
\newblock \href {https://doi.org/10.18653/v1/2023.emnlp-main.153} {G-{{Eval}}: {{NLG Evaluation}} using {{Gpt-4}} with {{Better Human Alignment}}}.
\newblock In \emph{Proceedings of the 2023 {{Conference}} on {{Empirical Methods}} in {{Natural Language Processing}}}, pages 2511--2522, {Singapore}. {Association for Computational Linguistics}.

\bibitem[{Min et~al.(2022)Min, Lyu, Holtzman, Artetxe, Lewis, Hajishirzi, and Zettlemoyer}]{minRethinkingRoleDemonstrations2022}
Sewon Min, Xinxi Lyu, Ari Holtzman, Mikel Artetxe, Mike Lewis, Hannaneh Hajishirzi, and Luke Zettlemoyer. 2022.
\newblock \href {https://doi.org/10.48550/arXiv.2202.12837} {Rethinking the {{Role}} of {{Demonstrations}}: {{What Makes In-Context Learning Work}}?}

\bibitem[{OpenAI et~al.(2023)OpenAI, Achiam, Adler, Agarwal, Ahmad, Akkaya, Aleman, Almeida, Altenschmidt, Altman, Anadkat, Avila, Babuschkin, Balaji, Balcom, Baltescu, Bao, Bavarian, Belgum, Bello, Berdine, {Bernadett-Shapiro}, Berner, Bogdonoff, Boiko, Boyd, Brakman, Brockman, Brooks, Brundage, Button, Cai, Campbell, Cann, Carey, Carlson, Carmichael, Chan, Chang, Chantzis, Chen, Chen, Chen, Chen, Chen, Chess, Cho, Chu, Chung, Cummings, Currier, Dai, Decareaux, Degry, Deutsch, Deville, Dhar, Dohan, Dowling, Dunning, Ecoffet, Eleti, Eloundou, Farhi, Fedus, Felix, Fishman, Forte, Fulford, Gao, Georges, Gibson, Goel, Gogineni, Goh, {Gontijo-Lopes}, Gordon, Grafstein, Gray, Greene, Gross, Gu, Guo, Hallacy, Han, Harris, He, Heaton, Heidecke, Hesse, Hickey, Hickey, Hoeschele, Houghton, Hsu, Hu, Hu, Huizinga, Jain, Jain, Jang, Jiang, Jiang, Jin, Jin, Jomoto, Jonn, Jun, Kaftan, Kaiser, Kamali, Kanitscheider, Keskar, Khan, Kilpatrick, Kim, Kim, Kim, Kirchner, Kiros, Knight, Kokotajlo, Kondraciuk, Kondrich, Konstantinidis, Kosic, Krueger, Kuo, Lampe, Lan, Lee, Leike, Leung, Levy, Li, Lim, Lin, Lin, Litwin, Lopez, Lowe, Lue, Makanju, Malfacini, Manning, Markov, Markovski, Martin, Mayer, Mayne, McGrew, McKinney, McLeavey, McMillan, McNeil, Medina, Mehta, Menick, Metz, Mishchenko, Mishkin, Monaco, Morikawa, Mossing, Mu, Murati, Murk, M{\'e}ly, Nair, Nakano, Nayak, Neelakantan, Ngo, Noh, Ouyang, O'Keefe, Pachocki, Paino, Palermo, Pantuliano, Parascandolo, Parish, Parparita, Passos, Pavlov, Peng, Perelman, Peres, Petrov, Pinto, Michael, Pokorny, Pokrass, Pong, Powell, Power, Power, Proehl, Puri, Radford, Rae, Ramesh, Raymond, Real, Rimbach, Ross, Rotsted, Roussez, Ryder, Saltarelli, Sanders, Santurkar, Sastry, Schmidt, Schnurr, Schulman, Selsam, Sheppard, Sherbakov, Shieh, Shoker, Shyam, Sidor, Sigler, Simens, Sitkin, Slama, Sohl, Sokolowsky, Song, Staudacher, Such, Summers, Sutskever, Tang, Tezak, Thompson, Tillet, Tootoonchian, Tseng, Tuggle, Turley, Tworek, Uribe, Vallone, Vijayvergiya, Voss, Wainwright, Wang, Wang, Wang, Ward, Wei, Weinmann, Welihinda, Welinder, Weng, Weng, Wiethoff, Willner, Winter, Wolrich, Wong, Workman, Wu, Wu, Wu, Xiao, Xu, Yoo, Yu, Yuan, Zaremba, Zellers, Zhang, Zhang, Zhao, Zheng, Zhuang, Zhuk, and Zoph}]{openai_gpt-4_2023}
OpenAI, Josh Achiam, Steven Adler, Sandhini Agarwal, Lama Ahmad, Ilge Akkaya, Florencia~Leoni Aleman, Diogo Almeida, Janko Altenschmidt, Sam Altman, Shyamal Anadkat, Red Avila, Igor Babuschkin, Suchir Balaji, Valerie Balcom, Paul Baltescu, Haiming Bao, Mo~Bavarian, Jeff Belgum, Irwan Bello, Jake Berdine, Gabriel {Bernadett-Shapiro}, Christopher Berner, Lenny Bogdonoff, Oleg Boiko, Madelaine Boyd, Anna-Luisa Brakman, Greg Brockman, Tim Brooks, Miles Brundage, Kevin Button, Trevor Cai, Rosie Campbell, Andrew Cann, Brittany Carey, Chelsea Carlson, Rory Carmichael, Brooke Chan, Che Chang, Fotis Chantzis, Derek Chen, Sully Chen, Ruby Chen, Jason Chen, Mark Chen, Ben Chess, Chester Cho, Casey Chu, Hyung~Won Chung, Dave Cummings, Jeremiah Currier, Yunxing Dai, Cory Decareaux, Thomas Degry, Noah Deutsch, Damien Deville, Arka Dhar, David Dohan, Steve Dowling, Sheila Dunning, Adrien Ecoffet, Atty Eleti, Tyna Eloundou, David Farhi, Liam Fedus, Niko Felix, Sim{\'o}n~Posada Fishman, Juston Forte, Isabella Fulford, Leo Gao, Elie Georges, Christian Gibson, Vik Goel, Tarun Gogineni, Gabriel Goh, Rapha {Gontijo-Lopes}, Jonathan Gordon, Morgan Grafstein, Scott Gray, Ryan Greene, Joshua Gross, Shixiang~Shane Gu, Yufei Guo, Chris Hallacy, Jesse Han, Jeff Harris, Yuchen He, Mike Heaton, Johannes Heidecke, Chris Hesse, Alan Hickey, Wade Hickey, Peter Hoeschele, Brandon Houghton, Kenny Hsu, Shengli Hu, Xin Hu, Joost Huizinga, Shantanu Jain, Shawn Jain, Joanne Jang, Angela Jiang, Roger Jiang, Haozhun Jin, Denny Jin, Shino Jomoto, Billie Jonn, Heewoo Jun, Tomer Kaftan, {\L}ukasz Kaiser, Ali Kamali, Ingmar Kanitscheider, Nitish~Shirish Keskar, Tabarak Khan, Logan Kilpatrick, Jong~Wook Kim, Christina Kim, Yongjik Kim, Hendrik Kirchner, Jamie Kiros, Matt Knight, Daniel Kokotajlo, {\L}ukasz Kondraciuk, Andrew Kondrich, Aris Konstantinidis, Kyle Kosic, Gretchen Krueger, Vishal Kuo, Michael Lampe, Ikai Lan, Teddy Lee, Jan Leike, Jade Leung, Daniel Levy, Chak~Ming Li, Rachel Lim, Molly Lin, Stephanie Lin, Mateusz Litwin, Theresa Lopez, Ryan Lowe, Patricia Lue, Anna Makanju, Kim Malfacini, Sam Manning, Todor Markov, Yaniv Markovski, Bianca Martin, Katie Mayer, Andrew Mayne, Bob McGrew, Scott~Mayer McKinney, Christine McLeavey, Paul McMillan, Jake McNeil, David Medina, Aalok Mehta, Jacob Menick, Luke Metz, Andrey Mishchenko, Pamela Mishkin, Vinnie Monaco, Evan Morikawa, Daniel Mossing, Tong Mu, Mira Murati, Oleg Murk, David M{\'e}ly, Ashvin Nair, Reiichiro Nakano, Rajeev Nayak, Arvind Neelakantan, Richard Ngo, Hyeonwoo Noh, Long Ouyang, Cullen O'Keefe, Jakub Pachocki, Alex Paino, Joe Palermo, Ashley Pantuliano, Giambattista Parascandolo, Joel Parish, Emy Parparita, Alex Passos, Mikhail Pavlov, Andrew Peng, Adam Perelman, Filipe de Avila~Belbute Peres, Michael Petrov, Henrique Ponde de~Oliveira Pinto, Michael, Pokorny, Michelle Pokrass, Vitchyr Pong, Tolly Powell, Alethea Power, Boris Power, Elizabeth Proehl, Raul Puri, Alec Radford, Jack Rae, Aditya Ramesh, Cameron Raymond, Francis Real, Kendra Rimbach, Carl Ross, Bob Rotsted, Henri Roussez, Nick Ryder, Mario Saltarelli, Ted Sanders, Shibani Santurkar, Girish Sastry, Heather Schmidt, David Schnurr, John Schulman, Daniel Selsam, Kyla Sheppard, Toki Sherbakov, Jessica Shieh, Sarah Shoker, Pranav Shyam, Szymon Sidor, Eric Sigler, Maddie Simens, Jordan Sitkin, Katarina Slama, Ian Sohl, Benjamin Sokolowsky, Yang Song, Natalie Staudacher, Felipe~Petroski Such, Natalie Summers, Ilya Sutskever, Jie Tang, Nikolas Tezak, Madeleine Thompson, Phil Tillet, Amin Tootoonchian, Elizabeth Tseng, Preston Tuggle, Nick Turley, Jerry Tworek, Juan Felipe~Cer{\'o}n Uribe, Andrea Vallone, Arun Vijayvergiya, Chelsea Voss, Carroll Wainwright, Justin~Jay Wang, Alvin Wang, Ben Wang, Jonathan Ward, Jason Wei, C.~J. Weinmann, Akila Welihinda, Peter Welinder, Jiayi Weng, Lilian Weng, Matt Wiethoff, Dave Willner, Clemens Winter, Samuel Wolrich, Hannah Wong, Lauren Workman, Sherwin Wu, Jeff Wu, Michael Wu, Kai Xiao, Tao Xu, Sarah Yoo, Kevin Yu, Qiming Yuan, Wojciech Zaremba, Rowan Zellers, Chong Zhang, Marvin Zhang, Shengjia Zhao, Tianhao Zheng, Juntang Zhuang, William Zhuk, and Barret Zoph. 2023.
\newblock \href {https://doi.org/10.48550/arXiv.2303.08774} {{{GPT-4 Technical Report}}}.

\bibitem[{Ouyang et~al.(2022)Ouyang, Wu, Jiang, Almeida, Wainwright, Mishkin, Zhang, Agarwal, Slama, Ray, Schulman, Hilton, Kelton, Miller, Simens, Askell, Welinder, Christiano, Leike, and Lowe}]{ouyang_training_2022}
Long Ouyang, Jeffrey Wu, Xu~Jiang, Diogo Almeida, Carroll Wainwright, Pamela Mishkin, Chong Zhang, Sandhini Agarwal, Katarina Slama, Alex Ray, John Schulman, Jacob Hilton, Fraser Kelton, Luke Miller, Maddie Simens, Amanda Askell, Peter Welinder, Paul~F. Christiano, Jan Leike, and Ryan Lowe. 2022.
\newblock \href {https://proceedings.neurips.cc/paper_files/paper/2022/hash/b1efde53be364a73914f58805a001731-Abstract-Conference.html} {Training language models to follow instructions with human feedback}.
\newblock \emph{Advances in Neural Information Processing Systems}, 35:27730--27744.

\bibitem[{Touvron et~al.(2023{\natexlab{a}})Touvron, Lavril, Izacard, Martinet, Lachaux, Lacroix, Rozi{\`e}re, Goyal, Hambro, Azhar et~al.}]{touvron2023llama}
Hugo Touvron, Thibaut Lavril, Gautier Izacard, Xavier Martinet, Marie-Anne Lachaux, Timoth{\'e}e Lacroix, Baptiste Rozi{\`e}re, Naman Goyal, Eric Hambro, Faisal Azhar, et~al. 2023{\natexlab{a}}.
\newblock Llama: Open and efficient foundation language models.
\newblock \emph{arXiv preprint arXiv:2302.13971}.

\bibitem[{Touvron et~al.(2023{\natexlab{b}})Touvron, Martin, Stone, Albert, Almahairi, Babaei, Bashlykov, Batra, Bhargava, Bhosale, Bikel, Blecher, Ferrer, Chen, Cucurull, Esiobu, Fernandes, Fu, Fu, Fuller, Gao, Goswami, Goyal, Hartshorn, Hosseini, Hou, Inan, Kardas, Kerkez, Khabsa, Kloumann, Korenev, Koura, Lachaux, Lavril, Lee, Liskovich, Lu, Mao, Martinet, Mihaylov, Mishra, Molybog, Nie, Poulton, Reizenstein, Rungta, Saladi, Schelten, Silva, Smith, Subramanian, Tan, Tang, Taylor, Williams, Kuan, Xu, Yan, Zarov, Zhang, Fan, Kambadur, Narang, Rodriguez, Stojnic, Edunov, and Scialom}]{touvron_llama_2023}
Hugo Touvron, Louis Martin, Kevin Stone, Peter Albert, Amjad Almahairi, Yasmine Babaei, Nikolay Bashlykov, Soumya Batra, Prajjwal Bhargava, Shruti Bhosale, Dan Bikel, Lukas Blecher, Cristian~Canton Ferrer, Moya Chen, Guillem Cucurull, David Esiobu, Jude Fernandes, Jeremy Fu, Wenyin Fu, Brian Fuller, Cynthia Gao, Vedanuj Goswami, Naman Goyal, Anthony Hartshorn, Saghar Hosseini, Rui Hou, Hakan Inan, Marcin Kardas, Viktor Kerkez, Madian Khabsa, Isabel Kloumann, Artem Korenev, Punit~Singh Koura, Marie-Anne Lachaux, Thibaut Lavril, Jenya Lee, Diana Liskovich, Yinghai Lu, Yuning Mao, Xavier Martinet, Todor Mihaylov, Pushkar Mishra, Igor Molybog, Yixin Nie, Andrew Poulton, Jeremy Reizenstein, Rashi Rungta, Kalyan Saladi, Alan Schelten, Ruan Silva, Eric~Michael Smith, Ranjan Subramanian, Xiaoqing~Ellen Tan, Binh Tang, Ross Taylor, Adina Williams, Jian~Xiang Kuan, Puxin Xu, Zheng Yan, Iliyan Zarov, Yuchen Zhang, Angela Fan, Melanie Kambadur, Sharan Narang, Aurelien Rodriguez, Robert Stojnic, Sergey Edunov, and Thomas Scialom. 2023{\natexlab{b}}.
\newblock \href {https://doi.org/10.48550/arXiv.2307.09288} {Llama 2: {{Open Foundation}} and {{Fine-Tuned Chat Models}}}.

\bibitem[{Wang et~al.(2023)Wang, Wang, Liu, Chen, Yuan, Peng, and Ji}]{wang_mint_2023}
Xingyao Wang, Zihan Wang, Jiateng Liu, Yangyi Chen, Lifan Yuan, Hao Peng, and Heng Ji. 2023.
\newblock \href {https://doi.org/10.48550/arXiv.2309.10691} {{{MINT}}: {{Evaluating LLMs}} in {{Multi-turn Interaction}} with {{Tools}} and {{Language Feedback}}}.

\bibitem[{Wei et~al.(2022)Wei, Wang, Schuurmans, Bosma, Ichter, Xia, Chi, Le, and Zhou}]{wei_chain--thought_2022}
Jason Wei, Xuezhi Wang, Dale Schuurmans, Maarten Bosma, Brian Ichter, Fei Xia, Ed~H. Chi, Quoc~V. Le, and Denny Zhou. 2022.
\newblock \href {https://openreview.net/forum?id=_VjQlMeSB_J} {Chain-of-{{Thought Prompting Elicits Reasoning}} in {{Large Language Models}}}.
\newblock In \emph{Advances in {{Neural Information Processing Systems}}}.

\bibitem[{Yao et~al.(2023)Yao, Yu, Zhao, Shafran, Griffiths, Cao, and Narasimhan}]{yao_tree_2023}
Shunyu Yao, Dian Yu, Jeffrey Zhao, Izhak Shafran, Thomas~L. Griffiths, Yuan Cao, and Karthik Narasimhan. 2023.
\newblock \href {https://doi.org/10.48550/arXiv.2305.10601} {Tree of {{Thoughts}}: {{Deliberate Problem Solving}} with {{Large Language Models}}}.

\bibitem[{Yue et~al.(2023)Yue, Ni, Zhang, Zheng, Liu, Zhang, Stevens, Jiang, Ren, Sun, Wei, Yu, Yuan, Sun, Yin, Zheng, Yang, Liu, Huang, Sun, Su, and Chen}]{yue_mmmu_2023}
Xiang Yue, Yuansheng Ni, Kai Zhang, Tianyu Zheng, Ruoqi Liu, Ge~Zhang, Samuel Stevens, Dongfu Jiang, Weiming Ren, Yuxuan Sun, Cong Wei, Botao Yu, Ruibin Yuan, Renliang Sun, Ming Yin, Boyuan Zheng, Zhenzhu Yang, Yibo Liu, Wenhao Huang, Huan Sun, Yu~Su, and Wenhu Chen. 2023.
\newblock \href {https://doi.org/10.48550/arXiv.2311.16502} {{{MMMU}}: {{A Massive Multi-discipline Multimodal Understanding}} and {{Reasoning Benchmark}} for {{Expert AGI}}}.

\bibitem[{Zhao et~al.(2023)Zhao, Zhou, Li, Tang, Wang, Hou, Min, Zhang, Zhang, Dong et~al.}]{zhao2023survey}
Wayne~Xin Zhao, Kun Zhou, Junyi Li, Tianyi Tang, Xiaolei Wang, Yupeng Hou, Yingqian Min, Beichen Zhang, Junjie Zhang, Zican Dong, et~al. 2023.
\newblock A survey of large language models.
\newblock \emph{arXiv preprint arXiv:2303.18223}.

\bibitem[{Zheng et~al.(2023{\natexlab{a}})Zheng, Chiang, Sheng, Li, Zhuang, Wu, Zhuang, Li, Lin, Xing, Gonzalez, Stoica, and Zhang}]{zheng_lmsys-chat-1m_2023}
Lianmin Zheng, Wei-Lin Chiang, Ying Sheng, Tianle Li, Siyuan Zhuang, Zhanghao Wu, Yonghao Zhuang, Zhuohan Li, Zi~Lin, Eric~P. Xing, Joseph~E. Gonzalez, Ion Stoica, and Hao Zhang. 2023{\natexlab{a}}.
\newblock \href {https://doi.org/10.48550/arXiv.2309.11998} {{{LMSYS-Chat-1M}}: {{A Large-Scale Real-World LLM Conversation Dataset}}}.

\bibitem[{Zheng et~al.(2023{\natexlab{b}})Zheng, Chiang, Sheng, Zhuang, Wu, Zhuang, Lin, Li, Li, Xing, Zhang, Gonzalez, and Stoica}]{zheng_judging_2023}
Lianmin Zheng, Wei-Lin Chiang, Ying Sheng, Siyuan Zhuang, Zhanghao Wu, Yonghao Zhuang, Zi~Lin, Zhuohan Li, Dacheng Li, Eric~P. Xing, Hao Zhang, Joseph~E. Gonzalez, and Ion Stoica. 2023{\natexlab{b}}.
\newblock \href {https://doi.org/10.48550/arXiv.2306.05685} {Judging {{LLM-as-a-Judge}} with {{MT-Bench}} and {{Chatbot Arena}}}.

\bibitem[{Zhou et~al.(2023)Zhou, Lu, Mishra, Brahma, Basu, Luan, Zhou, and Hou}]{zhou_instruction-following_2023}
Jeffrey Zhou, Tianjian Lu, Swaroop Mishra, Siddhartha Brahma, Sujoy Basu, Yi~Luan, Denny Zhou, and Le~Hou. 2023.
\newblock \href {https://doi.org/10.48550/arXiv.2311.07911} {Instruction-{{Following Evaluation}} for {{Large Language Models}}}.

\end{thebibliography}

\clearpage
\appendix

\newpage
\section{Prompts}
\label{sec:prompt}
This section details the specific prompts used to construct queries and evaluate responses in MT-Eval.

The prompt for adding additional turns in MT-Bench is provided in Figure \ref{fig:construct_mt-bench}.
The prompt for constructing information retrieval questions and context creation queries are depicted in Figure \ref{fig:construct_recollection_question} and \ref{fig:construct_recollection_prompt} respectively.
The prompt for creating the documents used in Refinement and Expansion tasks is outlined in Figure \ref{fig:construct_document}.
This section also includes prompts for converting multi-turn queries into a single-turn format for Follow-up and Refinement tasks, as shown in Figures \ref{fig:prompt_mt_bench_single} and \ref{fig:prompt_convert_single_inst}.
Finally, the evaluation of responses is guided by the chain-of-thought prompt in Figure \ref{fig:prompt_evaluation}.
\subsection{Data Construction Prompts}

\begin{strip}
  \begin{prompt}[width=\textwidth,label=fig:construct_mt-bench,breakable]{Prompt for creating additional turns in MT-Bench, which are used in the Follow-up task.}
    Formulate one extra conversation between a user and assistant where each user's question ideally links back to the previous relevant response from the assistant, fostering a layered and thoughtful conversation:\\
    \\
    \#\#\# Dialogue\\
    \{history\}\\
    \\
    \#\#\# Instruction\\
    In the following turns, the user's questions should aim to reference back to the previous assistant's responses, enhancing the interconnectedness of the dialogue while maintaining a coherent dialogue. Imagine if someone is reading your proposed question, he/she should not be able to understand it without reading the relevant Assistant's response. You should evolve the conversation around the topic of \{topic\}.\\
    \\
    \#\#\# Format\\
    You should output the following JSON:\\
    ```json\\
    \{\\
    "Rationale": <Explain how the user follows up the Assistant's response and which turn it is. Also, explain why it is related to \{topic\}>,\\
    "User": <The user's instruction or question.>,\\
    "Assistant": <The assistant's response.>\\
    \}\\
    ```
  \end{prompt}

  \begin{prompt}[width=\textwidth,label=fig:construct_recollection_question,breakable]{Prompt for creating information retrieval questions used in the Recollection task.}
    Propose 100 diverse questions in various domains. Domains include but are not limited to ethics, sports, music, art, science, literature, economics, medicine, food, technology, history, travel, and education. Questions can be open-ended or close-ended. Be creative!
  \end{prompt}
  \begin{prompt}[width=\textwidth,label=fig:construct_recollection_prompt,breakable]{Prompt for creating content creation queries used in Recollection task.}
    Propose 100 diverse content creation prompts in various domains. Domains include but are not limited to ethics, sports, music, art, science, literature, economics, medicine, food, technology, history, travel, and education. The content can be a product description, blog post, email, advertisement, story, pitch, speech, cover letter, etc. Be creative!
  \end{prompt}

  \begin{prompt}[width=\textwidth,label=fig:construct_document,breakable]{Prompt for creating documents of different media and topics. These documents are used in the Refinement and Expansion task.}
    \#\#\# Instruction\\
    Compose a hypothetical \{media\} about \{topic\} in about 250-300 words. Draw upon your creativity to feature people, locations, and objects that do not exist in history.\\
    \\
    You can follow the below steps to write:\\
    1. **Imaginative Elements**: Incorporate fictional characters, settings, and items. Make sure they are original and not based on real historical entities.\\
    \\
    2. **Outline**: Sketch a brief outline to organize your thoughts and plot points.\\
    \\
    3. **Write Your Draft**: Begin writing your piece, adhering to your outline and staying within the word limit.\\
    \\
    4. **Revise and Edit**: After your first draft, revise for clarity, creativity, and flow. Check your grammar and spelling.\\
    \\
    5. **Finalize**: Prepare the final version of your piece, ensuring it is polished and engaging.\\
    \\
    Provide only the final version in your response.\\
    \\
    \#\#\# Format\\
    Use the following format in your response:\\
    Topic: ...\\
    \{content\}\\
    \\
    \#\#\# Output
  \end{prompt}

  \begin{prompt}[width=\textwidth,label=fig:prompt_mt_bench_single,breakable,fontupper=\small]{The prompt to condense the relevant information in the dialogue history of the Follow-up task into one query.}
    Condense an user's question and the dialogue history between a user and an assistant into a single, concise question that includes all the necessary details without omitting any important information. The response should use the format "User: ..."\\
    \\
    \#\#\# Dialogue\\
    \{dialogue\}\\
    \\
    \#\#\# Question\\
    \{question\}
  \end{prompt}

  \begin{prompt}[width=\textwidth,label=fig:prompt_convert_single_inst,breakable,fontupper=\small]{The prompt to transform multiple instructions into one instruction for the single-turn Refinement task.}
    Below, I will provide you with a few instructions in a numbered list format. Your task is to condense these instructions into one coherent and concise instruction. Please note that if there are conflicting instructions later on, you should ignore the earlier conflicted constraints and prioritize the later ones. I want you to just output the condensed instruction without anything else. You should retain all the necessary elements from the original instructions.\\
    \\
    Instructions:\\
    \{constraint\}\\
    \\
    Now, condense the above instructions into one coherent and concise instruction. Provide your output in JSON format:\\
    \{\\
    "instruction": "<The condensed instruction.>"\\
    \}
  \end{prompt}

  \subsection{Evaluation Prompt}
  \begin{prompt}[width=\textwidth,label=fig:prompt_evaluation,breakable,fontupper=\small]{The evaluation prompt.}
    Evaluate the response provided below to determine if it meets the specified constraints related to the following article. Provide an integer score from 1 to 10, taking into account its helpfulness, relevance, accuracy, depth, creativity, and how well it conforms to the constraints. You should ignore any earlier constraints that contradict to the latter constraints. For constraints related to word and sentence counts, you must use my provided counts to judge whether the response fulfills the constraint. Before giving your score, you should first provide a rationale to explain it. \\
    \\
    Article to Evaluate Against:\\
    \{content\}\\
    \\
    Response to Evaluate:\\
    \{response\}\\
    \\
    Number of words in response: \{num\_words\} \\
    Number of sentences in response: \{num\_sent\} \\
    \\
    Constraints:\\
    \{constraints\}\\
    \\
    The evaluation must be structured in the following JSON format:\\
    ```json\\
    \{\\
    "Rationale": "<Explain the rationale of your score.>",\\
    "Score": "<An integer score from 1 to 10.>"\\
    \}\\
    ```
  \end{prompt}

\end{strip}

\null\newpage
\null\newpage
\section{Case Study}
\begin{strip}
  \begin{prompt}[width=\textwidth,label=fig:comma_inst,breakable]{An example case in the Recollection task showing all LLMs, except GPT-4, failed to refrain from using commas in the response.}
    \textcolor{red}{User:} All your responses in the upcoming conversation must refrain from the use of any commas. Keep all your responses under 200 words.\\
    \textcolor{red}{Assistant:} ok.\\
    \\
    \textcolor{red}{User:} Is the Mona Lisa Leonardo Da Vinci's most recognized piece?\\
    \textcolor{red}{GPT-4:} Yes the Mona Lisa is indeed the most recognized piece by Leonardo Da Vinci. It's a globally renowned artwork admired for its intricate detail and the mysterious expression of its subject.\\
    \textcolor{red}{Mixtral-Instruct-8x7B:} Yes, the Mona Lisa is Leonardo da Vinci's most recognized piece. It ...\\
    \textcolor{red}{Vicuna-13B-v1.5:} Yes, the Mona Lisa is Leonardo da Vinci's most recognized piece.\\
    \textcolor{red}{Qwen-chat-14B:} Yes, the Mona Lisa is Leonardo Da Vinci's most recognized piece.
  \end{prompt}

  \begin{prompt}[width=\textwidth,label=fig:case_refine,breakable]{An example case in the Refinement task where the model adheres to the latest instructions while neglecting earlier ones. Here, the model correctly paraphrases the third paragraph with future perfect tense and starts every sentence with the letter 'S'. However, it fails to incorporate a metaphor and alliteration into each sentence. The sentences are also longer than 10 words.}
    \textcolor{red}{User:} Content: ... \\
    \\
    Instruction: Please rewrite or paraphrase the third paragraph from the initially provided content, ensuring the new version maintains the original meaning but uses different wording. Do not include any further explanations or extra output in your response.  \\
    \textcolor{red}{Assistant:} The third paragraph rewritten is: "Star players of the match were Jaxon 'Lightning' Vega from the Solar Flares, admired for his swift maneuvers and accurate shots, and Lyra 'Meteor' Kael, the captain of Nebula Knights, celebrated for her tactical decisions and exceptional defense skills. The game was a thrilling exhibition of prowess, strategy, and cutting-edge technology."\\
    \\
    \textcolor{red}{User:} Use at least one metaphor in your rewrite.  \\
    \textcolor{red}{Assistant:} The third paragraph rewritten with a metaphor is: "Jaxon 'Lightning' Vega of the Solar Flares and Lyra 'Meteor' Kael, captain of the Nebula Knights, shone as the match's metaphorical sun and moon, with Vega's nimble moves and pinpoint shooting contrasting Kael's strategic brilliance and unyielding defense. Their contrasting styles set the stage for a gripping contest of skill, strategy, and futuristic technology."\\
    \\
    \textcolor{red}{User:} Now, include each sentence to be no more than 10 words.  \\
    \textcolor{red}{Assistant:} Jaxon 'Lightning' Vega (Solar Flares) and Lyra 'Meteor' Kael (Nebula Knights) starred.\\
    Vega's swift moves and accurate shots contrasted Kael's defense and strategy.\\
    A thrilling display of skill, strategy, and futuristic tech unfolded.\\
    \\
    \textcolor{red}{User:} Next, incorporate alliteration into each sentence.  \\
    \textcolor{red}{Assistant:} Swiftly soaring, Jaxon "Lightning" Vega led the Luminous Lunar Flares.\\
    Lyra "Meteor" Kael, the Nebula Knights' noble navigator, countered.\\
    A breathtaking ballet of basketball, soccer, and sci-fi technology transpired. \\
    \\
    \textcolor{red}{User:} This time, write it in future perfect tense.  \\
    \textcolor{red}{Assistant:} By the end of the championship, Jaxon "Lightning" Vega will have demonstrated his remarkable agility and precision.\\
    Lyra "Meteor" Kael will have orchestrated her team's defense and strategy to counter the opposition.\\
    The spectators will have witnessed a captivating contest of skill, strategy, and futuristic technology. \\
    \\
    \textcolor{red}{User:} Finally, start every sentence with the letter 'S'.\\
    \textcolor{red}{Assistant:} Spectators will be enthralled by the nimble moves of Jaxon "Lightning" Vega, a Solar Flares player.\\
    Simultaneously, Lyra "Meteor" Kael, captain of the Nebula Knights, will command attention with her strategic prowess and defensive skills.\\
    Showcasing a harmonious blend of basketball, soccer, and state-of-the-art hover technology, the match will unfold.
  \end{prompt}

  \begin{prompt}[width=\textwidth,label=fig:eval_wrong,breakable,fontupper=\small]{An example showing GPT-4 failure to identify relevant constraints which result in inaccurate evaluation.}
    Evaluate the response provided below to determine if it meets the specified constraints related to the following article. Provide an integer score from 1 to 10, taking into account its helpfulness, relevance, accuracy, depth, creativity, and how well it conforms to the constraints. You should ignore any earlier constraints that contradict to the latter constraints. For constraints related to word and sentence counts, you must use my provided counts to judge whether the response fulfills the constraint. Before giving your score, you should first provide a rationale to explain it. \\
    \\
    \textcolor{red}{Article to Evaluate Against:}\\
    Dear Dr. Elara Thorne,\\
    \\
    I hope this message finds you in good health. I am reaching out to discuss the ethical implications surrounding our newest endeavor at Orion Innovations - the MindWeaver. Your expertise in the realm of technology ethics is essential for guiding our strategy in this groundbreaking project.\\
    \\
    ...\\
    \\
    \textcolor{blue}{Response to Evaluate:}\\
    Our ethical endeavor utilizes expertise in Orion Innovations' MindWeaver technology.\\
    \\
    Number of words in response: 10 \\
    Number of sentences in response: 1 \\
    \\
    \textcolor{orange}{Constraints:}\\
    1. List all the adjectives and nouns in the first paragraph initially provided content. Your answers should be separated by commas and formatted as 'Adjectives: ..., ...; Nouns: ..., ...'.\\
    2. Now, list only those from the first three sentences of the initially provided content.\\
    3. Among these, list those begin with the letters 'e', 'm', or 'o'.\\
    4. Arrange in alphabetical order\\
    5. Compose a coherent sentence using all the identified adjectives and nouns in your previous response.\\
    6. Your coherent sentence should be within 20 words.\\
    \\
    The evaluation must be structured in the following JSON format:\\
    ```json\\
    \{\\
    "Rationale": "<Explain the rationale of your score.>",\\
    "Score": "<An integer score from 1 to 10.>"\\
    \}\\
    ```\\
    \\
    \multilinecolortext{magenta}{
      Evaluation Response: \\
      \{\\
      "Rationale": "The response does not meet any of the constraints provided. It does not list any adjectives or nouns, it does not focus on the first three sentences of the initial content, it does not identify words beginning with 'e', 'm', or 'o', it does not arrange anything in alphabetical order, and it does not compose a coherent sentence using identified adjectives and nouns. The response is also not relevant to the article's content.",\\
      "Score": "1"\\
      \}\\
    }
  \end{prompt}
\end{strip}

\null\newpage
\null\newpage
\section{Task Examples}
\begin{strip}
  \begin{prompt}[width=\textwidth,label=fig:refinement_ex,breakable,fontupper=\small]{An example of the Refinement task.}
    \textcolor{red}{User:} Content: In the bustling city of Gastropolis, where the scent of spices and the sizzle of pans wafted through the air, there lived a chef named Luminara, whose culinary skills were the stuff of legend. Luminara’s restaurant, The Enchanted Pot, was tucked away in an alley draped with ivy and fairy lights, a hidden gem known only to the most adventurous of food lovers.\\
    \\
    Luminara was not just any chef; she was a culinary alchemist, blending traditional ingredients with rare, magical flora and fauna. Her signature dish was the Whimsical Soup, a broth that changed flavor with the mood of the diner, leaving them delighted and astonished in equal measure.\\
    \\
    The story begins on a crisp autumn morning when Luminara received a mysterious letter sealed with an emerald wax stamp. The letter was an invitation to embark on a quest to discover the legendary Moonberry, an elusive ingredient said to unlock the true potential of any dish it graced. It was an adventure that could cement Luminara's place in the annals of culinary history.\\
    \\
    With her trusty talking spatula, Spatulon, by her side, Luminara set out on her quest. She traversed through the Whispering Woods, where trees shared secrets and the air shimmered with enchantment. She scaled the peaks of Mount Savor, each step bringing her closer to the Moonberry, which was guarded by the mythical creature known as the Gastrogriff.\\
    \\
    Upon reaching the peak at twilight, Luminara found the Gastrogriff perched beside a single, luminescent Moonberry bush. With a respectful nod to the majestic beast, she approached and explained her quest for culinary greatness. The Gastrogriff, impressed by Luminara's passion and determination, offered her a single Moonberry, its glow reflecting in her hopeful eyes.\\
    \\
    Luminara returned to Gastropolis, her apron stained with the adventures of her journey. The Moonberry was the star of her next creation, a dish that didn't just resonate with the diner's mood but also told a story, a story of a chef's quest for the extraordinary.\\
    \\
    As patrons of The Enchanted Pot took their first bites, they were transported through Luminara's journey, tasting the whispering woods, the icy peaks, and the warmth of triumph. Luminara had not just found an ingredient; she had woven her tale into the tapestry of Gastropolis's rich culinary lore. The Enchanted Pot was no longer just a restaurant; it was a portal to the wonders of imagination, one dish at a time. \\
    \\
    Instruction: Write a short summary based on the initially provided content. Do not include any further explanations or extra output in your response.\\
    \textcolor{red}{User:} Focus only on Luminara's quest and her interaction with the Gastrogriff.\\
    \textcolor{red}{User:} Use at most 30 words.\\
    \textcolor{red}{User:} Rewrite it using Luminara's first-person perspective.\\
    \textcolor{red}{User:} In the summary, use metaphors to describe Luminara's quest and encounter with the Gastrogriff.\\
    \textcolor{red}{User:} Avoid using the words 'Luminara', 'Moonberry', and 'Gastrogriff'.
  \end{prompt}

  \begin{prompt}[width=\textwidth,label=fig:expansion_ex,breakable,fontupper=\small]{An example of the Expansion task.}
    \textcolor{red}{User:} Content: In the bustling city of Gastropolis, where the scent of spices and the sizzle of pans wafted through the air, there lived a chef named Luminara, whose culinary skills were the stuff of legend. Luminara’s restaurant, The Enchanted Pot, was tucked away in an alley draped with ivy and fairy lights, a hidden gem known only to the most adventurous of food lovers.\\
    \\
    Luminara was not just any chef; she was a culinary alchemist, blending traditional ingredients with rare, magical flora and fauna. Her signature dish was the Whimsical Soup, a broth that changed flavor with the mood of the diner, leaving them delighted and astonished in equal measure.\\
    \\
    The story begins on a crisp autumn morning when Luminara received a mysterious letter sealed with an emerald wax stamp. The letter was an invitation to embark on a quest to discover the legendary Moonberry, an elusive ingredient said to unlock the true potential of any dish it graced. It was an adventure that could cement Luminara's place in the annals of culinary history.\\
    \\
    With her trusty talking spatula, Spatulon, by her side, Luminara set out on her quest. She traversed through the Whispering Woods, where trees shared secrets and the air shimmered with enchantment. She scaled the peaks of Mount Savor, each step bringing her closer to the Moonberry, which was guarded by the mythical creature known as the Gastrogriff.\\
    \\
    Upon reaching the peak at twilight, Luminara found the Gastrogriff perched beside a single, luminescent Moonberry bush. With a respectful nod to the majestic beast, she approached and explained her quest for culinary greatness. The Gastrogriff, impressed by Luminara's passion and determination, offered her a single Moonberry, its glow reflecting in her hopeful eyes.\\
    \\
    Luminara returned to Gastropolis, her apron stained with the adventures of her journey. The Moonberry was the star of her next creation, a dish that didn't just resonate with the diner's mood but also told a story, a story of a chef's quest for the extraordinary.\\
    \\
    As patrons of The Enchanted Pot took their first bites, they were transported through Luminara's journey, tasting the whispering woods, the icy peaks, and the warmth of triumph. Luminara had not just found an ingredient; she had woven her tale into the tapestry of Gastropolis's rich culinary lore. The Enchanted Pot was no longer just a restaurant; it was a portal to the wonders of imagination, one dish at a time. \\
    \\
    Instruction: Translate the first paragraph to Chinese. Just provide the translation directly without any further explanations or extra output.\\
    \textcolor{red}{User:} Base on the initially provided content, answer the question: What magical ingredient did Luminara seek on her quest, and which mythical creature guarded it?\\
    User: Write a short summary based on the initially provided content. Do not include any further explanations or extra output in your response.\\
    \textcolor{red}{User:} List all the relations of the types {[based in, work for, located in, live in]} among the entities {[person, location, organization]} in the initially given content. Just provide the relations that were explicitly stated in the context without any further explanations or extra output. Provide the relations in the format of (entity 1, relation, entity 2), (entity 1, relation, entity 2), .... For example: (Shi Liming, work for, Institute of Zoology).\\
    \textcolor{red}{User:} List all the persons and places in the initially provided content. Your answers should be separated by commas and formatted as 'Person: ..., ...; Places: ..., ...'.\\
    \textcolor{red}{User:} List all the adjectives in the initially provided content. Your answers should be separated by commas. Do not include any further explanations or extra output in your response.\\
    \textcolor{red}{User:} Classify the initially provided content into one of the following labels: ethics, sports, music, art, science, literature, economics, medicine, food, technology, history, travel, education. Just provide the correct label without any further explanations or extra output.
  \end{prompt}

  \begin{prompt}[width=\textwidth,label=fig:follow_up_ex,breakable,fontupper=\small]{An example of the Follow-up task. The first two turns are from MT-Bench \citep{zheng_judging_2023}.}
    \textcolor{red}{User:} Thomas is very healthy, but he has to go to the hospital every day. What could be the reasons?\\
    \textcolor{red}{User:} Can you explain why the above question is interesting?\\
    \textcolor{red}{User:} I'm curious about the possibility of therapy or rehabilitation being the reason. Can you give me some examples of injuries or conditions that might require daily therapy sessions?\\
    \textcolor{red}{User:} If Thomas is indeed undergoing daily therapy, what would be some signs or behaviors that might indicate this? I'm thinking of things beyond just visiting the hospital.\\
    \textcolor{red}{User:} So, if Thomas exhibits multiple signs from the list you provided, like using a walking aid and experiencing chronic pain, it's highly likely he's undergoing daily therapy?
  \end{prompt}

  \begin{prompt}[width=\textwidth,label=fig:recollection_ex,breakable,fontupper=\small]{An example of the Recollection task that tests the model's obedience to a global instruction given in the beginning.}
    \textcolor{red}{User:} Begin all your responses in the upcoming conversation with the letter o. Keep all your responses under 200 words.\\
    \textcolor{red}{User:} Is it morally wrong to break a law you consider unjust?\\
    \textcolor{red}{User:} What role do sports play in promoting unity and cultural understanding?\\
    \textcolor{red}{User:} Write a product description for a paint set that can help beginner artists hone their skills.\\
    \textcolor{red}{User:} Is a vegan diet healthier than a diet including meat?\\
    \textcolor{red}{User:} What impact does deforestation have on our global climate?\\
    \textcolor{red}{User:} Write a blog post talking about tips to ace job interview.
  \end{prompt}
\end{strip}

\clearpage
\null\newpage
\null\newpage

\section{Human Evaluation}
\label{sec:human_evaluation}
We recruit five graduate students to evaluate 60 randomly selected instances from each of the Follow-up, Refinement, and Expansion tasks.
To ensure consistency with the GPT-4 evaluation, we provided the same instructions used for GPT-4 (Figure \ref{fig:prompt_evaluation}).
Before the main evaluation, we measured the inter-rater reliability of the students by having them evaluate another 20 random instances.
The Cohen's kappa score of 0.58 indicated satisfactory agreement.

\section{Benchmark Construction Details}
\label{sec:benchmark_detail}

This section provides the details of constructing Recollection, Follow-up, Refinement, and Expansion tasks.
\paragraph{Recollection}
We construct two Recollection subtasks of varying difficulty.  The first is a document classification task, where the model categorizes a document into one of the 13 categories, namely ethics, sports, music, art, science, literature, economics, medicine, food, technology, history, travel, and education.
We present 100 documents across 10 dialogues, each with 10 turns.
The documents are generated using GPT-4, which covers nine document types, including news articles, blog posts, emails, posters, social media posts, book chapters, research papers, advertisements, and legal documents.
The second subtask involves following a global instruction provided in the initial turn throughout the dialogue.
We construct this task based on IFEval \citep{zhou_instruction-following_2023}. At each subsequent turn,  the user poses content creation or information retrieval questions.
These questions are generated by GPT-4 (the full prompts are available in Figure \ref{fig:construct_recollection_question} and \ref{fig:construct_recollection_prompt}).
We select 14 varied instructions from IFEval that align well with content creation and information retrieval tasks to use as the initial instruction. For each instruction, we construct two dialogue sessions with 10 dialogue turns, resulting in 280 dialogue turns.
The description of the instructions used can be found in Table \ref{tab:inst_desc}.

The first task is simpler as models can refer to the dialogue context to understand the task and identify the labels used for classification, whereas the second requires models to consistently recall the initial instruction, which is more challenging.
An example of this task can be found in Figure \ref{fig:recollection_ex}.

\paragraph{Follow-up}
To construct the Follow-up dialogues, we expand the 80 two-turn dialogues from MT-Bench \citep{zheng_judging_2023} by adding three extra turns, adding 240 dialogue turns in total.
To expand the dialogues, we first employ GPT-4 to generate a preliminary follow-up question and a corresponding answer (the complete prompt is provided in Figure \ref{fig:construct_mt-bench}). Then, we recruit student helpers to review and refine the content as necessary. An illustrative example of a follow-up task is presented in Figure \ref{fig:follow_up_ex}.

\paragraph{Refinement}
We construct the Refinement tasks by formulating two document-based NLP tasks within a dialogue instance.
Each NLP task consists of six instructions of increasing complexity.
These instructions are generated by prompting GPT-4 to produce five queries for each of the four NLP tasks: question-answering, summarization, named-entity recognition, and paragraph rewriting.
We guide GPT-4 to impose an additional constraint on each successive query (the full prompt is available in Figure \ref{fig:construct_document}).
We create four dialogue instances per document, with each of the four NLP tasks serving as the initial task, and a different task as the second task.
This process results in 40 dialogue instances, corresponding to 480 dialogue turns.
An example refinement task can be found in Figure \ref{fig:refinement_ex}.

For the single-turn scenario, we utilize GPT-4 to convert the multiple instructions into a single instruction using the prompt detailed in Figure \ref{fig:prompt_convert_single_inst}.

\paragraph{Expansion}
The Expansion task challenges the model with a series of NLP tasks based on the initially provided document.
Each instance consists of seven NLP tasks in random order, including question-answering, summarization, named-entity recognition, part-of-speech tagging, relation extraction, translation, and classification.
Figure \ref{fig:expansion_ex} illustrates an example of this task.

\begin{table*}[h]
  \centering

  \begin{tabular}{ll}
    \toprule
    Instruction                           & Description                                      \\
    \midrule
    startend:start\_char                  & Start with a particular letter.                  \\
    startend:start\_emoji                 & Start with an emoji.                             \\
    startend:end\_phrase                  & End with a particular phrase.                    \\
    language:response\_language           & Respond in a particular language.                \\
    format:json\_format                   & Respond using JSON.                              \\
    format:bullet\_list                   & Using a specified number of bullet lists.        \\
    length\_constraints:number\_sentences & Respond with a specified sentence limit.         \\
    keywords:existence                    & Include some specific keywords.                  \\
    change\_case:capital\_letter          & Respond in uppercase.                            \\
    change\_case:lowercase                & Respond in lowercase.                            \\
    punctuation:no\_comma                 & Refrain from using commas.                       \\
    combination:two\_responses            & Include two different responses.                 \\
    content:placeholder                   & Include a certain amount of placeholders.        \\
    format:constrained\_response          & Reply with one of the provided response options. \\
    \bottomrule
  \end{tabular}

  \caption{The descriptions of the instructions used in the Recollection task. Most of it is adopted from \citet{zhou_instruction-following_2023}.}
  \label{tab:inst_desc}

\end{table*}

\section{Additional Statistics}

\begin{table*}[htbp]
  \centering
  \begin{tabular}{c|rrrrr}
    \toprule
                          & Recollection            & Expansion               & Refinement              & Follow-up               & All                     \\
    \midrule
    Avg. \# Turns         & \multirow{2}{*}{10}     & \multirow{2}{*}{7.00}   & \multirow{2}{*}{12.00}  & \multirow{2}{*}{3.00}   & \multirow{2}{*}{6.96}   \\
    per Dialogue          &                         &                         &                         &                         &                         \\
    \hline
    Avg. \# Words         & \multirow{2}{*}{693.09} & \multirow{2}{*}{539.60} & \multirow{2}{*}{882.85} & \multirow{2}{*}{686.82} & \multirow{2}{*}{760.41} \\
    in Prompt $\dagger$   &                         &                         &                         &                         &                         \\
    \hline
    Max. \# Words         & \multirow{2}{*}{2331}   & \multirow{2}{*}{838}    & \multirow{2}{*}{2574}   & \multirow{2}{*}{1932}   & \multirow{2}{*}{2574}   \\
    in Prompt $\dagger$   &                         &                         &                         &                         &                         \\
    \hline
    Avg. \# Words         & \multirow{2}{*}{72.07}  & \multirow{2}{*}{24.41}  & \multirow{2}{*}{78.50}  & \multirow{2}{*}{205.88} & \multirow{2}{*}{99.31}  \\
    in Response $\dagger$ &                         &                         &                         &                         &                         \\
    \hline
    Max. \# Words         & \multirow{2}{*}{289}    & \multirow{2}{*}{107}    & \multirow{2}{*}{430}    & \multirow{2}{*}{444}    & \multirow{2}{*}{444}    \\
    in Response $\dagger$ &                         &                         &                         &                         &                         \\
    \hline
    Avg. \# Words         & \multirow{2}{*}{54.49}  & \multirow{2}{*}{156.77} & \multirow{2}{*}{65.89}  & \multirow{2}{*}{31.78}  & \multirow{2}{*}{60.63}  \\
    per Turn              &                         &                         &                         &                         &                         \\
    \hline
    Max. \# Words         & \multirow{2}{*}{330}    & \multirow{2}{*}{474}    & \multirow{2}{*}{449}    & \multirow{2}{*}{262}    & \multirow{2}{*}{474}    \\
    per Turn              &                         &                         &                         &                         &                         \\
    \hline
    Total \# Dialogues    & \multirow{1}{*}{38}     & \multirow{1}{*}{10}     & \multirow{1}{*}{40}     & \multirow{1}{*}{80}     & \multirow{1}{*}{168}    \\
    \hline
    Total \# Turns        & \multirow{1}{*}{380}    & \multirow{1}{*}{70}     & \multirow{1}{*}{480}    & \multirow{1}{*}{240}    & \multirow{1}{*}{1170}   \\
    \bottomrule
  \end{tabular}
  \caption{Detailed statistics for individual tasks in \name. $\dagger$: Estimated using GPT-4 responses.}
  \label{tab:detaiiled_stat}
\end{table*}
We show the statistics of each task in Table \ref{tab:detaiiled_stat}.

\section{Additional Results}
We show the breakdown of the results in the Recollect task in Table \ref{tab:breakdown_rec}.
\begin{table*}[h]
  \centering
  \begin{tabular}{lrrrr}
    \toprule
    \multirow{2}{*}{Model} & \multicolumn{2}{c}{Global Inst.} & \multicolumn{2}{c}{CLS}                                       \\
    \cmidrule(lr){2-3} \cmidrule(lr){4-5}
                           & Single-Turn                      & Multi-Turn              & Single-Turn      & Multi-Turn       \\
    \midrule
    GPT-3.5-Turbo          & 8.00                             & 4.61                    & \highlight{9.50} & 9.20             \\
    GPT-4                  & \highlight{9.75}                 & \highlight{9.71}        & 9.30             & \highlight{9.50} \\
    ChatGLM3-6B            & 3.61                             & 2.61                    & 6.50             & 3.24             \\
    Vicuna-7B-v1.5         & 4.61                             & 3.11                    & 8.10             & 7.80             \\
    Vicuna-13B-v1.5        & 6.46                             & 4.64                    & 7.50             & 7.90             \\
    Llama-2-chat-7B        & 6.43                             & 2.32                    & 8.10             & 5.40             \\
    Llama-2-chat-13B       & 7.32                             & 2.93                    & 7.70             & 4.40             \\
    Qwen-chat-7B           & 6.04                             & 2.89                    & 8.30             & 7.60             \\
    Qwen-chat-14B          & 7.39                             & 4.43                    & 9.40             & 8.00             \\
    Mistral-Instruct-7B    & 7.54                             & 5.25                    & 9.40             & 9.19             \\
    Mixtral-Instruct-8x7B  & 6.93                             & 3.21                    & 8.80             & 9.13             \\
    \bottomrule
  \end{tabular}
  \caption{Breakdown of results in the Recollection task. Global Inst. stands for the global instruction following task, and CLS denotes the document classification task.}
  \label{tab:breakdown_rec}
\end{table*}

\section{The Impact of Dialogue History in the Follow-up Task.}
\label{sec:ablation_followup}
\begin{table*}[h]
  \centering
  \resizebox{\textwidth}{!}{
    \begin{tabular}{lrrrrr}
      \toprule
      Instruction                           & GPT-4 & Vicuna-13B-v1.5 & Qwen-chat-14B & Mistral-Instruct-7B & Mixtral-Instruct-8x7B \\
      \midrule
      change\_case:capital\_letter          & 10.00 & 1.50            & 0.50          & 3.50                & 0.50                  \\
      change\_case:lowercase                & 10.00 & 0.00            & 2.00          & 2.50                & 1.50                  \\
      combination:two\_responses            & 10.00 & 0.00            & 4.50          & 0.00                & 3.50                  \\
      content:placeholder                   & 5.00  & 1.50            & 1.50          & 5.50                & 1.50                  \\
      format:bullet\_list                   & 9.00  & 6.00            & 6.00          & 2.00                & 6.00                  \\
      format:constrained\_response          & 10.00 & 10.00           & 10.00         & 5.00                & 5.00                  \\
      format:json\_format                   & 10.00 & 8.00            & 8.00          & 8.00                & 3.00                  \\
      keywords:existence                    & 10.00 & 0.00            & 0.00          & 2.00                & 1.50                  \\
      language:response\_language           & 10.00 & 0.00            & 9.50          & 9.50                & 2.50                  \\
      length\_constraints:number\_sentences & 6.50  & 5.00            & 3.00          & 0.00                & 0.00                  \\
      punctuation:no\_comma                 & 10.00 & 0.00            & 0.00          & 0.00                & 0.00                  \\
      startend:end\_phrase                  & 6.00  & 9.50            & 6.00          & 10.00               & 6.00                  \\
      startend:start\_char                  & 10.00 & 5.00            & 5.00          & 1.50                & 0.00                  \\
      startend:start\_emoji                 & 10.00 & 7.00            & 0.50          & 6.50                & 0.00                  \\
      \bottomrule
    \end{tabular}
  }
  \caption{The number of turns that different models can adhere to the global instructions in the Recollection task, averaged over two dialogues per instruction.}
  \label{tab:turns_in_recollection}
\end{table*}

\begin{table*}[h]
  \centering
  \begin{tabular}{lrr}
    \toprule
    Model                 & Single-Turn Avg. & Multi-Turn Avg.  \\
    \midrule
    GPT-3.5-Turbo         & 9.19             & 9.21             \\
    GPT-4                 & \highlight{9.24} & \highlight{9.60} \\
    ChatGLM3-6B           & 7.60             & 8.39             \\
    Vicuna-7B-v1.5        & 7.88             & 8.31             \\
    Vicuna-13B-v1.5       & 8.37             & 8.68             \\
    Llama-2-chat-7B       & 7.97             & 8.53             \\
    Llama-2-chat-13B      & 8.30             & 8.82             \\
    Qwen-chat-7B          & 7.98             & 8.49             \\
    Qwen-chat-14B         & 8.60             & 9.12             \\
    Mistral-Instruct-7B   & 8.78             & 9.05             \\
    Mixtral-Instruct-8x7B & 9.02             & 9.52             \\
    \bottomrule
  \end{tabular}
  \caption{Performance of single-turn and multi-turn setting in the Follow-up task. The result of multi-turn is exacted from Table \ref{tab:multi_turn_main_table}.}
  \label{tab:follow_up}
\end{table*}

We conduct further study to explore the role of dialogue history as in-context examples and its impact on model performance in the Follow-up task.
In this task, the user's query directly follows the assistant's previous reply.
Modifying the dialogue history is not as straightforward as in the document classification task outlined in Section \ref{sec:ablation}.
To address this, we convert the multi-turn setting into a single-turn one, presenting only a complete query without prior responses.
We employ GPT-4 to condense the relevant information from the dialogue history into a query for all 240 instances, using the prompt provided in Figure \ref{fig:prompt_mt_bench_single}.
However, the initial queries often omit crucial contextual information. Consequently, we meticulously review and refine each generated query as needed.

Table \ref{tab:follow_up} presents the performance of various models in both single-turn and multi-turn settings.
All models exhibit significantly better performance in the multi-turn setting, indicating the positive impact of dialogue history on model performance.
To gain further insights, we analyze 30 instances and compare the differences in the two settings.
We find that the models often leverage previous responses and explanations to generate improved responses in the current turn.
This also suggests that high-quality dialogue history plays the role of in-context learning examples, providing useful guidance to the model's response.  Figure \ref{fig:case_follow_up} shows an example with Qwen-chat-7b, illustrating how the multi-turn response benefits from the dialogue history's intermediate reasoning, leading to an accurate current-turn response.

\begin{prompt}[float*,width=\textwidth,label=fig:case_follow_up,breakable,fontupper=\small]{Comparison of the responses in single-turn and multi-turn setting of an instance in the Follow-up task.}
  \textcolor{red}{Single-Turn Prompt:} \\
  Consider this two scenarios:\\
  1. David has three sisters. Each of them has one brother.\\
  2. David has three sisters. Each of them has two brother.\\
  \\
  Can you explain how you determined the number of brothers David has in each scenario using logical reasoning? \\
  \\
  Single-Turn Response: In scenario 1, David has three sisters, and each of them has one brother. Therefore, David has a total of 3 sisters * 1 brother/sister = 3 brothers.\\
  \\
  In scenario 2, David has three sisters, and each of them has two brothers. Therefore, David has a total of 3 sisters * 2 brothers/sister = 6 brothers.\\
  \\
  In both scenarios, the number of brothers David has is determined by multiplying the number of sisters he has by the number of brothers each sister has.\\
  \\
  \textcolor{red}{Multi-Turn Prompt:} \\
  User: David has three sisters. Each of them has one brother. How many brothers does David have?\\
  Assistant: David has only one brother.\\
  User: If we change the previous question and assume that each sister of David has two brothers, how many brothers would David have?\\
  Assistant: If each of David's sisters has two brothers, then David would have two brothers.\\
  User: I have some doubts regarding your first answer. If David's sisters have one brother, and David is also the brother of his sister, then how could David has one brother?\\
  Assistant: I apologize for any confusion caused by my previous response. Let's clarify:\\
  \\
  In the original scenario where David has three sisters and each of them has one brother, David is indeed that one brother. Therefore, David has no other brothers; he is the only brother to his three sisters.\\
  \\
  In the modified scenario where each of David's sisters has two brothers, it implies that David has one additional brother besides himself. Therefore, in this case, David would have one brother.\\
  User: Can you explain how you determined the number of brothers David has in each scenario using logical reasoning?\\
  \\
  \textcolor{red}{Multi-Turn Response:} \\
  Sure, I'd be happy to explain my reasoning.\\
  \\
  In the original scenario where David has three sisters and each of them has one brother, we can use logical reasoning to determine that David has no other brothers. Here's how:\\
  \\
  1. David has three sisters.\\
  2. Each of David's sisters has one brother.\\
  3. Therefore, David is the only brother to his three sisters.\\
  \\
  In the modified scenario where each of David's sisters has two brothers, we can use logical reasoning to determine that David has one additional brother besides himself. Here's how:\\
  \\
  1. David has three sisters.\\
  2. Each of David's sisters has two brothers.\\
  3. Therefore, David has one additional brother besides himself.\\
  4. Since David has one additional brother, he has a total of two brothers.\\
  \\
  I hope this helps clarify my reasoning. Let me know if you have any further questions.
\end{prompt}

\end{document}